\documentclass[runningheads]{llncs} 
\usepackage{graphicx}
\graphicspath{{images/}}
\usepackage{amsmath}
\usepackage{amssymb}
\usepackage{xcolor}

\usepackage{graphicx}
\usepackage{nicefrac}  
\usepackage{setspace} 

\usepackage{multirow}

\usepackage{hyperref}
\hypersetup{
    colorlinks=true,
    linkcolor=blue,  
    citecolor=blue,   
    urlcolor=blue,
}

\usepackage[ruled,vlined]{algorithm2e}
\usepackage{algorithmic}

\newcommand{\dotp}[2]{\langle #1,\,#2\rangle}
\newcommand{\norm}[1]{|\!| #1 |\!|}

\DeclareMathOperator{\Prox}{Prox}
\DeclareMathOperator{\Proj}{Proj}

\newcommand{\R}{\mathbb{R}}
\newcommand{\V}{\mathcal{V}}
\newcommand{\E}{\mathcal{E}}
\newcommand{\G}{\mathcal{G}}
\newcommand{\N}{\mathcal{N}}

\newcommand{\ie}{\emph{i.e.}~}
\newcommand{\eg}{\emph{e.g.}~}

\newlength{\mylength}

\begin{document}
\title{
Variational models for signal processing\\ with Graph Neural Networks
\thanks{This paper has been accepted for \emph{International Conference on Scale Space and Variational Methods in Computer Vision} (SSVM-2021) .}
} 

%

\author{
Amitoz  Azad\inst{} \and
Julien  Rabin\inst{} \and
Abderrahim Elmoataz\inst{}
}

\authorrunning{A. Azad et al.}

\institute{
Normandie Univ, UNICAEN, ENSICAEN, CNRS, GREYC, 14000 Caen, France\\
\email{firstname.lastname@unicaen.fr}
}

\maketitle

\begin{abstract}
This paper is devoted to signal processing on point-clouds by means of neural networks. Nowadays, state-of-the-art in image processing and computer vision is mostly based on training deep convolutional neural networks on large datasets. While it is also the case for the processing of point-clouds with Graph Neural Networks (GNN), the focus has been largely given to high-level tasks such as classification and segmentation using supervised learning on labeled datasets such as ShapeNet.
Yet, such datasets are scarce and time-consuming to build depending on the target application. In this work, we investigate the use of variational models for such GNN to process signals on graphs for unsupervised learning.

Our contributions are two-fold. 
We first show that some existing variational - based algorithms for signals on graphs can be formulated as Message Passing Networks (MPN), a particular instance of GNN, making them computationally efficient in practice when compared to standard gradient-based machine learning algorithms. 
Secondly, we investigate the unsupervised learning of feed-forward GNN, either by
direct optimization of an inverse problem or by model distillation from variational-based MPN.

\end{abstract}

\section{Introduction}
\label{intro}
Variational methods have been a popular framework for signal processing since the last few decades. 
Although, with the advent of deep Convolutional Neural Networks (CNN), most computer vision tasks and image processing problems 
are nowadays addressed using state-of-the-art machine learning techniques.
Recently, there has been a growing interest in hybrid approaches combining machine learning techniques with variational modeling have seen a (see \eg \cite{meinhardt2017learning,kobler2017variational,bertocchi2020deep,combettes2020deep,hasannasab2020parseval}).

\subsubsection{Motivations.}
In this paper, we focus on such hybrid approaches for processing of point-clouds and signals on graphs.
Graph representation provides a unified framework to work with regular and irregular data, such as images, text, sound, manifold, social network and arbitrary high dimensional data. 
Recently, many machine learning approaches have also been proposed to deal with high-level computer vision tasks, such as classification and segmentation~\cite{qi2016pointnet,qi2017pointnetplusplus,wang2019dynamic}, point-cloud generation~\cite{yang2019pointflow}, surface reconstruction \cite{williams2019deep}, point cloud registration \cite{wang2019deep} and up-sampling \cite{li2019pu} 
to name a few. 
These approaches are built upon Graph Neural Networks (GNN) which are inspired by powerful techniques employed in CNN, such as Multi-Linear Perceptron (MLP), pooling, convolution, \emph{etc}. 
The main caveat is that transposing CNN architectures for graph-like inputs is not a simple task. Compared to images which are defined on a standard, regular grid that can be easily sampled, data with graph structures can have a large structural variability, which makes the design of a universal architecture difficult.
This is likely the main reason why 
GNN are still hardly being used for signal processing on graph for which variational methods are still very popular. 
In this work, we investigate the use of data-driven machine learning techniques combined with variational models to solve inverse problems on point clouds.

\subsection{Related work}

\subsubsection{Graph Neural Networks}
GNN are  types of Neural Networks which directly operate on the graph structure. 
GNN are emerging as a strong tool to do representation learning on various types of data which, similarly to CNN, can hierarchically learn complex features.

PointNet~\cite{qi2016pointnet} was one of the first few successful attempts to build GNN. Rather than preprocessing the input point-cloud by casting it into a generic structure, Qi et al. proposed to tackle the problem of structure irregularity by only considering permutation invariant operators that are applied to each vertex independently.
Further works such as \cite{qi2017pointnetplusplus} have built upon this framework to replicate other successful NN techniques, such as pooling and interpolation in auto-encoder, which enables to process neighborhood of points. 

Most successful attempts to build Graph \emph{Convolutional} Networks (GCN) were based upon graph spectral theory (see for instance \cite{bruna2014spectral,defferrard2016convolutional,kipf2017semi}).
In particular, Kipf and Welling~\cite{kipf2017semi} proposed a simple, linear propagation rule which can be seen as a first-order approximation of local spectral filters on graphs.
More recently, Wang et al. introduced an edge convolution technique~\cite{wang2019dynamic} which maintains the permutation invariance property.
Interestingly, \cite{pmlr-v97-wu19e} 
showed that GCN may act as low pass filters on internal features.

\subsubsection{Message Passing Networks}
At the core of the most the aforementioned GNNs, is \emph{Message Passing} (MP). The term was first coined in~\cite{gilmer2017neural}, in which authors proposed a common framework that reformulated many existing GNNs, such as \cite{bruna2014spectral,qi2016pointnet,kipf2017semi}. 
In the simplest terms, MP is a generalized convolution operator on irregular domains which typically expresses some neighborhood aggregation mechanism consisting of a message function, a permutation invariant function (\eg $\max$) and an update function.

\subsubsection{Inverse problems on Graphs.}
PDEs and variational methods are two major frameworks
to address inverse problems on a regular grid structure like images. These methods have also been extended to weighted graphs in non-local form by defining $p$-Laplace operators~\cite{elmoataz2015p}. 
Typical but not limited applications of these operators are filtering, classification, segmentation, or inpainting~\cite{peyre2011non}. 
Specifically, we are interested in this work in non-local Total-Variation on a graph, which has been thoroughly studied, \eg for non-local signal processing~\cite{elmoataz2008nonlocal,gilboa2009nonlocal,tabti2018symmetric},
multi-scale decomposition of signals~\cite{hidane2013nonlinear},
point-cloud segmentation~\cite{lozes2013nonlocal} and sparsification~\cite{tenbrinck2019variational},
using various optimization schemes 
(such as finite difference schemes~\cite{elmoataz2008nonlocal},
primal-dual~\cite{chambolle2011first},
forward-backward~\cite{raguet2015preconditioning},
or cut pursuit~\cite{raguet2018cut}).

\subsubsection{Contributions and Outline}

The paper is organised as follows. 
Section~\ref{sec:Inverse_Problem_MPN} focus on variational methods to solve inverse problems for graph signal processing.
After a short overview of graph representation and non-local regularization, 
we demonstrate that two popular optimization algorithms can be interpreted as a specific instance of Message Passing Networks (MPNs). 
This allows using efficient GPU-based machine learning libraries to solve the inverse problem on graphs. %
Experiments on point cloud compression and color processing show that such {MPN optimization is more efficient than} standard machine learning optimization techniques.
Section~\ref{sec:GNN} is devoted to the use of a trainable GNN to perform signal processing on graphs in a \emph{feed-forward} fashion. 
We investigate two techniques for unsupervised training, \ie in absence of ground-truth data, using variational models as prior knowledge to drive the optimization.
We show that either model distillation from a variational MPN or optimizing the inverse problem directly with GNN gives satisfactory approximate solutions and allows for faster processing.
\section{Solving Inverse Problem with MPN optimization}
\label{sec:Inverse_Problem_MPN}
\subsection{Notations and definitions} 
\label{sec:notations}
A graph $\G = (\V, \E, \omega)$ is composed of an ensemble of nodes (vertices) $\V$, an ensemble of edges $\E = \V \times \V$ and weights {$\omega : \E \mapsto \R_{+}$}, 
{where $\R_{+}$ indicates the set of non-negative values $[0,\infty)$}.
Let $f_{i} \in \R^{d}$ represents the feature vector (signal) on the node $i$ of $\G$, 
$w_{i,j}$ the scalar weight on the edge from the node $i$ to $j$, 
and $j \in \N(i)$ indicates a node $j$ in the neighborhood of node $i$, such that $\omega_{i,j} \neq 0$. 
For sake of simplicity, we only consider non-directed graph, \ie defined with symmetric weights $\omega_{i,j} = \omega_{j,i}$.

We refer to the weighted difference operator as $\nabla_\omega$:
    $ 
        (\nabla_\omega f)_{i,j}  = \sqrt{\omega_{i,j}} \left(f_j - f_i\right) \in \R^d
    $
    ,
    ${\forall\ (i,j) \in \E}  
    $. 
Assuming symmetry of $\omega$, the adjoint of this linear operator is 
    $
        (\nabla^\ast_\omega g)_{i} = \sum_{j \in \cal V}  \sqrt{\omega_{i,j}}\left(g_{j,i}-g_{i,j}\right) \in  \R^d,
        \forall\ i \in \V.
    $
We denote as $\|.\|_p$ the $\ell_{p}$ norm and $\| . \|_{1,p}$ 
the composed norm for parameter $p \in \R_+ \cup \{\infty\}$: 
\begin{equation*}
    \forall\ g \in \R^{|\V|  \times |\V| \times d}, \;
    \| g \|_{1,p} 
        = \sum_{i \in \V} \| g_{i,.} \|_p 
        = \sum_{i \in \V} \Big( \sum_{j \in \V,1 \le k \le d} |g_{i,j,k}|^p \Big)^{1/p} 
    .
\end{equation*}

\subsection{Non-Local Regularization on Graph}
In this work, we focus on inverse problems for signal processing on graph using the non-local regularization framework \cite{elmoataz2008nonlocal}. 
The optimization problem is formulated generically using $\ell_2$ fidelity term and a non-local regularization defined from the $\ell^p$ norm, parametrized by weights $\omega$ and power coefficient $q \in \R_+$, and penalized by $\lambda \in \R_+$ 
\begin{equation}\label{eq:optimization_problem}
    \inf_f \left\{{J}(f) := \tfrac{1}{2}\|f-f_0\|^2 + \tfrac{\lambda}{q} \text{R}_{p}^q(f)
    \right\}
    \text{ where } \text{R}_{p}^q(f) = {\|\nabla_\omega f\|}_{1,p}^q
    .
\end{equation}
Here $f_0$ is the given signal which requires processing.
In experiments, we will mainly focus on two cases:
when $p=q=2$, the smooth regularization term relates to the Tikhonov regularization and boils down to Laplacian diffusion on a graph \cite{elmoataz2008nonlocal}.
For $q=1$, the regularization term is the Non-Local Total Variation (NL-TV), referred to as \emph{isotropic} when $p=2$ and \emph{anisotropic} when $p=1$.
Other choice of norm might be useful: using $p\leq1$, as studied for instance for $p=0$ in~\cite{tenbrinck2019variational}, results in sparsification of signals;
using $p=\infty$~\cite{tabti2018symmetric} is also useful when considering unbiased symmetric schemes.

As the functional $J$ is not smooth, an $\varepsilon$-approximation $J_\varepsilon$ is often considered to circumvent numerical problems (see \eg \cite{elmoataz2008nonlocal}); this is for instance achieved by substituting the term $R_p^q$ with
(respectively for the case $p=2$ and $p=1$)
\begin{equation}\label{eq:NL-TV_eps}
    \text{R}^q_{2,\varepsilon}(f) = \sum_{i \in \V} \left( \|\nabla_\omega f_{i,.}\|_p^2 + \varepsilon^2\right)^{\frac{q}{2}}
    \text{ and }
    \text{R}^q_{1,\varepsilon}(f) = \sum_{i,j \in \V}  \omega_{i,j}^{\frac{q}{2}} \left(|f_{i} - f_{j}| + \varepsilon\right)^q
    .
\end{equation}

\subsection{Variational Optimization} 

We consider now two popular algorithms to solve inverse problem \eqref{eq:optimization_problem}, described in Alg.~\ref{algo:GJ} and \ref{algo:CP} with update rules on $f^{(t)}$, where $(t)$ is the iteration number.

\noindent\hspace*{-1mm}
\begin{minipage}{.5\textwidth}
\begin{algorithm}[H]\caption{Gauss-Jacobi 
\cite{elmoataz2008nonlocal}}\label{algo:GJ}
\small
\begin{algorithmic}
\STATE Initialization: $f^{(0)} = f_0$, 
set $\varepsilon>0$
\STATE compute $\gamma^{(t)}_{i,j}$ using Eq.\eqref{eq:gamma_update_GJ}
\STATE $\bar f^{(t+1)}_i = (f_0)_i + \lambda \sum_{j\in \N_{i}} \gamma^{(t)}_{i,j} f^{(t)}_{j}$ \\
\STATE $g^{(t+1)}_i = 1 + \lambda \sum_{j\in \N_{i}} \gamma^{(t)}_{i,j} $\\
\STATE $f^{(t+1)}_i = {\bar f^{(t+1)}_i} / {g^{(t+1)}_i}$ \\
\end{algorithmic}
\end{algorithm}
\end{minipage}
~\vrule~
\begin{minipage}{.49\textwidth}
\setstretch{1.25}
\begin{algorithm}[H]\caption{Primal-Dual 
\cite{chambolle2011first}
}\label{algo:CP}
\small
\begin{algorithmic}
\STATE Parameters: $\tau, \beta >0$ and $\theta \in [0,1]$\\
\STATE Initialization: 
$f^{(0)} = \bar{f}^{(0)} = f_0$\\ 
\STATE $g^{(t+1)} = \Prox_{\beta A^\ast }(g^{(t)}+\beta K \bar{f}^{(t)})$\\
\STATE $f^{(t+1)} = \Prox_{\tau B}(f^{(t)}-\tau K^\ast g^{(t+1)})$\\
\STATE $\bar{f}^{(t+1)}= f^{(t+1)}+ \theta(f^{(t+1)}-f^{(t)})$
\end{algorithmic}
\end{algorithm}
\end{minipage}

For any $p,q>0$, one can turn to the Gauss-Jacobi (GS) iterated filter \cite{elmoataz2008nonlocal} described in Alg.~\ref{algo:GJ}.
It relies on $\varepsilon$ approximation to derive the objective function, giving for instance the following update rule when using Eq.~\eqref{eq:NL-TV_eps}
\begin{equation}\label{eq:gamma_update_GJ}
    \gamma^{(t)}_{i,j} = 
    \begin{cases}
        w_{i,j} \left( 
        (\|(\nabla_{w}f)_{i,.}\|^2 + \varepsilon^2)^\frac{q-2}{2} + (\|(\nabla_{w}f)_{j,.}\|^2 + \varepsilon^2)^\frac{q-2}{2} 
        \right)
        & \text{if } p=2
        \\
        2 {w_{i,j}}^{\frac{q}{2}} \left( |f_{i} - f_{j}| + \varepsilon\right)^{q-2}
        & \text{if } p=1
    \end{cases}
    .
\end{equation}

When considering the specific case of NL-TV ($q=1$), one can turn to the primal-dual algorithm \cite{chambolle2011first} described in Alg.~\ref{algo:CP}, as done in \cite{lozes2013nonlocal,hidane2013nonlinear}. 
To do so, we recast the problem \eqref{eq:optimization_problem} as
$
\min_f \max_g \dotp{Kf}{g} - A^\ast(g) + B(f)
$
where $K \equiv \nabla_\omega$, 
$A\equiv {\| . \|}_{1,p}$, 
$B \equiv \tfrac{1}{2\lambda}{\|.-f_0\|}_2^2$.
Parameters must be chosen such that $ \tau \beta \norm{K}^2<1$ to ensure convergence. For numerical experiments, we have used $\theta=1$ and $\tau = \beta = (4 \max_{i \in \V} \sum_{j \in \V} \omega_{i,j})^{-1}$.
Note that other methods might be used, such as pre-conditioning \cite{raguet2015preconditioning}, and cut pursuit \cite{raguet2018cut}. 
Proximal operators corresponds to 
\begin{equation}
\Prox_{\tau B}(f) = \tfrac{\lambda f + \tau f_0}{\lambda + \tau}
\;,
\quad \text{ and } 
\Prox_{\beta A^*}(g) = \Proj_{\mathcal B_{\infty,p'}}(g)
\end{equation}
where the dual norm parameter $p'>0$ verifies $1/p' + 1/p = 1$.
The projection on the unit ball $\mathcal B_{\infty,p'}$ for $p=2$ and $p=1$ is given respectively by, $\forall \ g \in \R^{|\V| \times |\V| \times d}$
\begin{equation}\label{eq:proj}
\Proj_{\mathcal B_{\infty,2}}(g)_{i,j,k}
= \frac{g_{i,j,k}}{\max \{1, {\|g_{i,.}\|}_2\}}
\;,
\Proj_{\mathcal B_{\infty,\infty}}(g)_{i,j,k}
= \frac{g_{i,j,k}}{\max \{1, | g_{i,j,k} |\}}
.
\end{equation}

\subsection{Message Passing Network Optimization} 
\label{MPN}

Message Passing Networks~\cite{gilmer2017neural} can be formulated as follows
\begin{equation}
\label{eq:messagepassing} 
    {f}_i^{(n+1)}
        =\psi^{(n)} \left( {f}_i^{(n)}, \square{}_{j \in \N(i)}\  \phi^{(n)}\left({f}_i^{(n)}, {f}_j^{(n)},\omega_{i,j}\right) \right)
    \quad
    \forall \ i \in \V,\; 
\end{equation}
where $(n)$ indicates the depth in the network.
As already shown in~\cite{gilmer2017neural}, many existing GNN can be recast in this generic framework. 
Typically, $\square{}$ is the summation operator but can be any differentiable permutation invariant functions (such as $\max$, $\text{mean}$); $\psi^{(n)}$ (update rule) and $\phi^{(n)}$ (message rule) are the differentiable operators, such as MLPs (\ie affine functions combined with simple non-linear functions such as RELU).
Note that they are independent of the input vertex index to preserve permutation invariance.

We now show that Alg.~\ref{algo:GJ} and \ref{algo:CP}
can be also formulated within this framework.
\paragraph{G-J algorithm.}
    This is quite straightforward for the anisotropic case ($p=1$) using two MPNs to compute the update of $f$ in Alg.~\ref{algo:GJ}.
    A first MPN is used for the numerator $\bar f$, identifying $\psi^{(n)}(f_i^{(n)},F_i^{(n)}) = (f_0)_i + \lambda F_i^{(n)}$,
    $\square{} = \sum$ and 
    $\phi^{(n)}(f_i^{(n)}, f_j^{(n)}, \omega_{ij}) = \gamma_{i,j}^{(n)} f_j^{(n)}$ where $\gamma_{i,j}^{(n)}$, as defined in Eq.~\eqref{eq:gamma_update_GJ}, is a function of the triplet $(f_i^{(n)}, f_j^{(n)}, \omega_{i,j})$.
    A second MPN is used to compute the denominator $g$ using $\psi^{(n)}(f_i^{(n)},F_i^{(n)}) = 1 + \lambda F_i^{(n)}$ and $\phi^{(n)}(f_i^{(n)}, f_j^{(n)}, \omega_{ij}) = \gamma_{i,j}^{(n)}$.
    For the isotropic case $p=2$, an additional MPN is ultimately required to compute the norm of the gradient $(Kf^{(n)})_i$ at each vertex $i$, as detailed in the next paragraph. 
    
\paragraph{P-D algorithm.}
    Two \emph{nested} MPNs are now required to compute the update of the primal variable $f$ and the dual variable $g$ in Alg.~\ref{algo:CP}.
    For $f$, the MPN reads as the affine operator {$\psi^{(n)}(f_i^{(n)}, {F}_i^{(n)} ) = \tfrac{\tau}{\tau+\lambda} (f_0)_i + \tfrac{\tau}{\tau+\lambda} (f_i^{(n)} - \tau {F}_i^{(n)})$.} 
    Similarly, for the update of $g$ one computes the projection:
    {$\psi^{(n)}(g_i^{(n)}, {G}_i^{(n)} )  = \Proj_{\mathcal B_{\infty,p'}}(g_i^{(n)}+\beta {G}_i^{(n)}).$}
    The gradient operator $K$ and its adjoint $K^*$ (see definition in \S.~\ref{sec:notations}) can be applied using the processing functions $\phi$ and $\phi_j$, respectively
    {
    $$ 
    {G}_i^{(n)} 
        = (K \bar f^{(n)})_i 
        = \phi^{(n)} (\bar f_i^{(n)}, \bar f_j^{(n)},  \omega_{i,j})
        = \omega_{i,j} (\bar f_j^{(n)}- \bar f_i^{(n)}) 
    $$ 
   } 
   {
    $$
        {F}_i^{(n)} 
            = (K^* g^{(n+1)}))_i 
            = \square{} \phi_j^{(n)} (g_{i}^{(n+1)}, g_{j}^{(n+1)},  \omega_{i,j})
            = \sum_{j\in {N}_{i}} \omega_{i,j} (g_{j,i}^{(n+1)}- g_{i,j}^{(n+1)}) 
    $$
   } 
Observe that $\phi_j$ is a slightly modified message passing function using index $j$ to incorporate the edge features from not only source $i$ to target $j$ (\ie $g_{i,j}$ and $\omega_{i,j}$) but also from target $j$ to source $i$ ($g_{j,i}$ and $\omega_{j,i}$). 

These MPN formulations of Alg.~\ref{algo:GJ} and Alg.~\ref{algo:CP} allows to use machine learning libraries to solve the inverse problem Eq.~\ref{eq:optimization_problem} and to take full advantage of fast GPU optimization. 
In the following experiments, we illustrate the advantage of such MPN optimization in comparison with standard gradient-based machine learning optimization techniques relying on auto-differentiation.

\subsection{Experiments on point-clouds and comparison with auto-diff}

\paragraph{Experimental setting.} 
Edges on point-clouds are defined with weights $\omega_{i,j} \in \{0,1\}$ using the indicator function of a k-nearest neighbor search ($k=4$) on points coordinates using $\ell_2$ norm, then imposing symmetry by setting $\omega_{i,j} \leftarrow \omega_{i,j} \vee \omega_{j,i}$. 
Two types of features $f$ 
are tested for $q=1$, as illustrated in Fig.~\ref{fig:loss evolution}: 3D point coordinates $x$ for point-cloud simplification (Fig.~\ref{fig:elephant_inverse}), and colors $c \in [0,1]^{|\V| \times 3}$ for denoising (Fig.~\ref{fig:napoleon_inverse}). 
We also consider the non-convex case where $q=0.1$ for point cloud sparsification in Fig.~\ref{fig:elephant_inverse}.

\paragraph{Auto-differentiation.} As a baseline, we compare these MPN-based algorithms with algorithmic auto-differentiation. In this setting, the update of features at iteration $(t)$ is defined by gradient descent on the loss function $J(f^{(t)})$ defined in Eq.~\eqref{eq:optimization_problem}, \eg $f^{(t+1)} = f^{(t)} - \rho^{(t)} D^{(t)}$.
Since the NL-TV regularization term is not smooth, we use the $\varepsilon$ approximation of Eq.~\eqref{eq:NL-TV_eps} to compute the gradient $D^{(t)}$.
We consider here various standard gradient descent techniques used for NN training: Gradient Descent (\textsc{gd}), \textsc{adam} \cite{kingma2014adam} and \textsc{lbfgs}~\cite{liu1989limited}.

\paragraph{Implementation details.} 
All algorithms are implemented using the \emph{Pytorch Geometric} library~\cite{torchgeo} and tested with a Nvidia GPU GTX1080Ti with 11 GB of memory. 
Note that reported computation time is only indicative, as implementation greatly influence the efficiency of memory and GPU cores allocation.
Different learning rates are used for each method in the experiments.
The learning rate of \textsc{adam} algorithm is always set to $lr = 0.001$. \textsc{gd} is also used with $lr = 0.001$ for point-cloud simplification but with $lr = 0.1$ for color processing.
\textsc{lbfgs} is set with $lr = 0.01$ for $q=1$, 
and with $lr = 0.1$ for $q=0.1$.

\setlength{\mylength}{0.24\linewidth}
\setlength{\tabcolsep}{1pt}
\renewcommand{\arraystretch}{1.}
\begin{figure}[htb]
    \centering
    \begin{tabular}{c ccc}     
        Features $f_0$
        & $q=1$, $p=2$
        & $q=1$, $p=1$
        & $q=0.1$, $p=2$
        \\
        \includegraphics[width=\mylength]{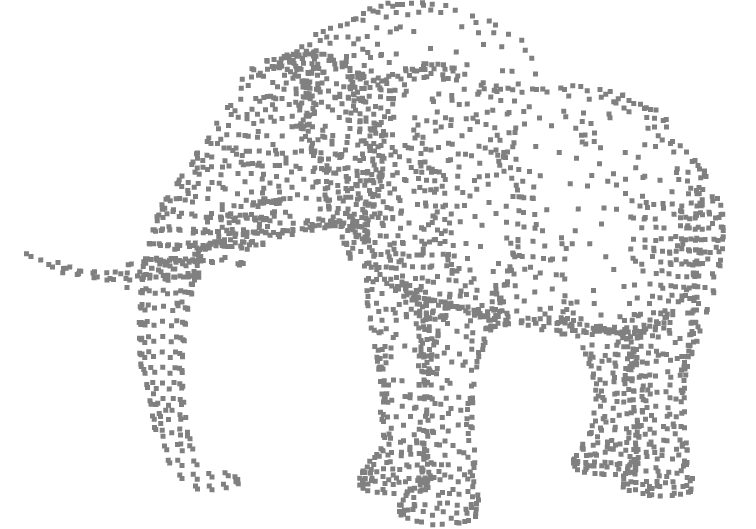} 
        &\includegraphics[width=\mylength]{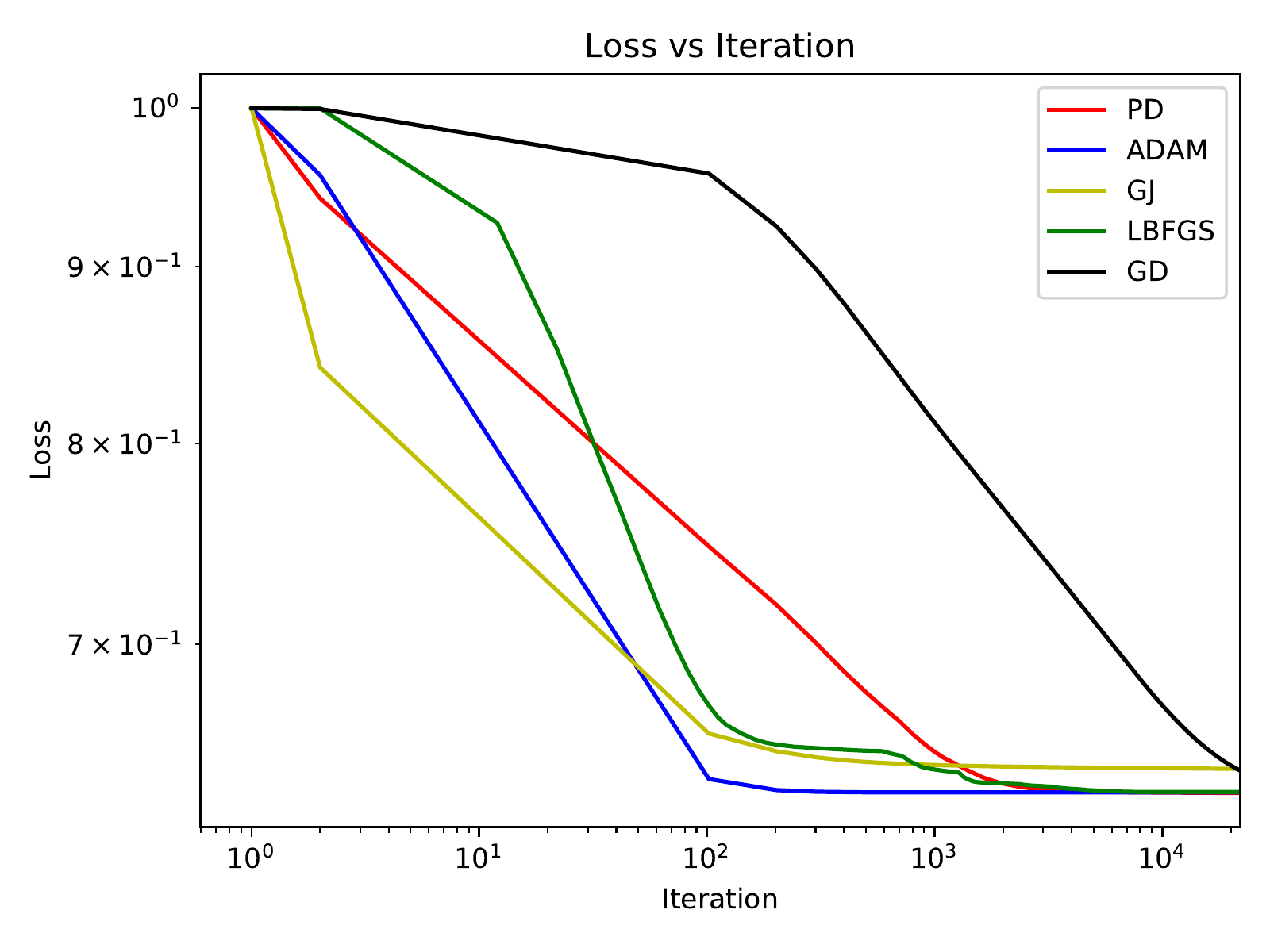}
        & \includegraphics[width=\mylength]{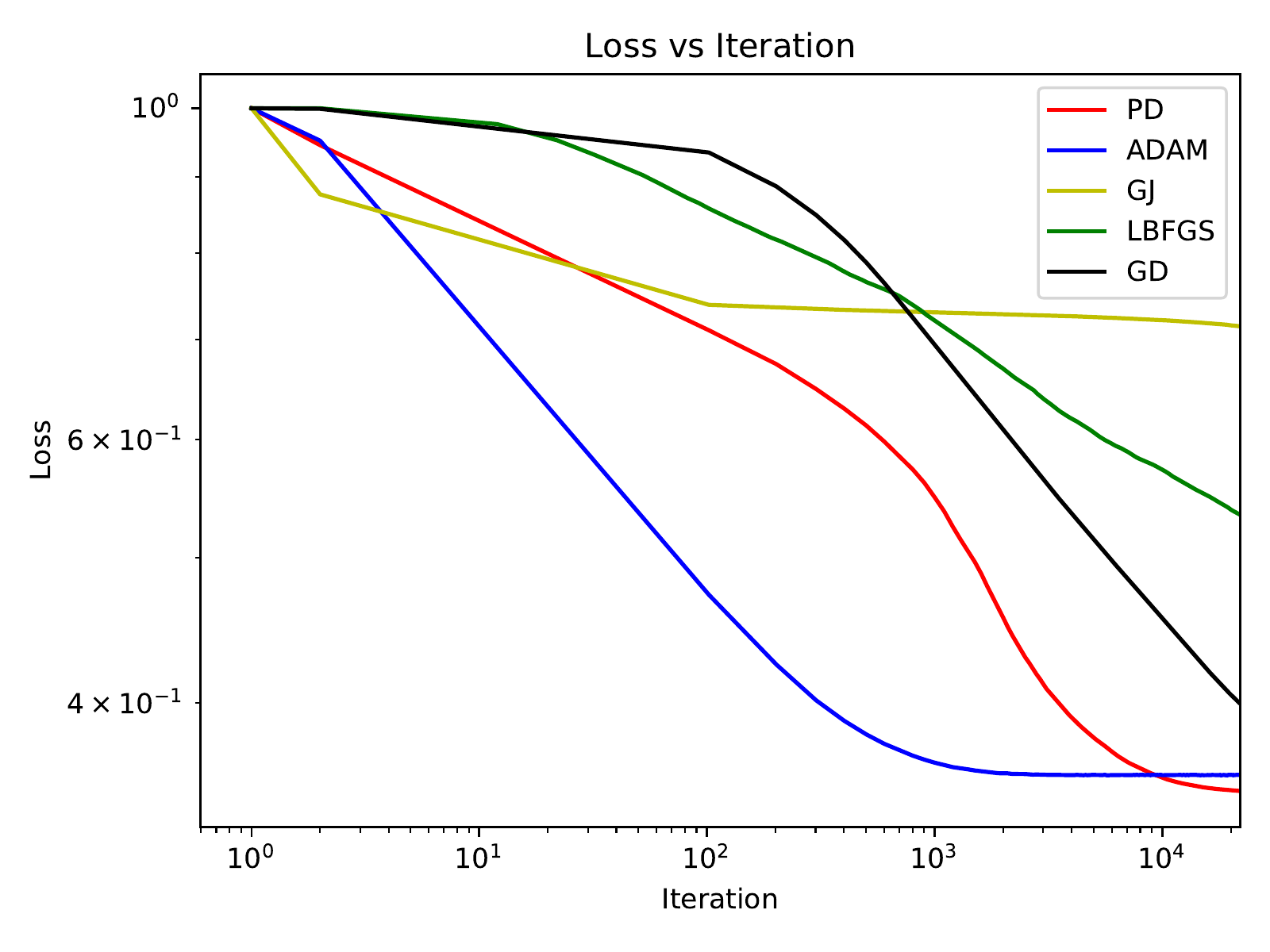}
        & \includegraphics[width=\mylength]{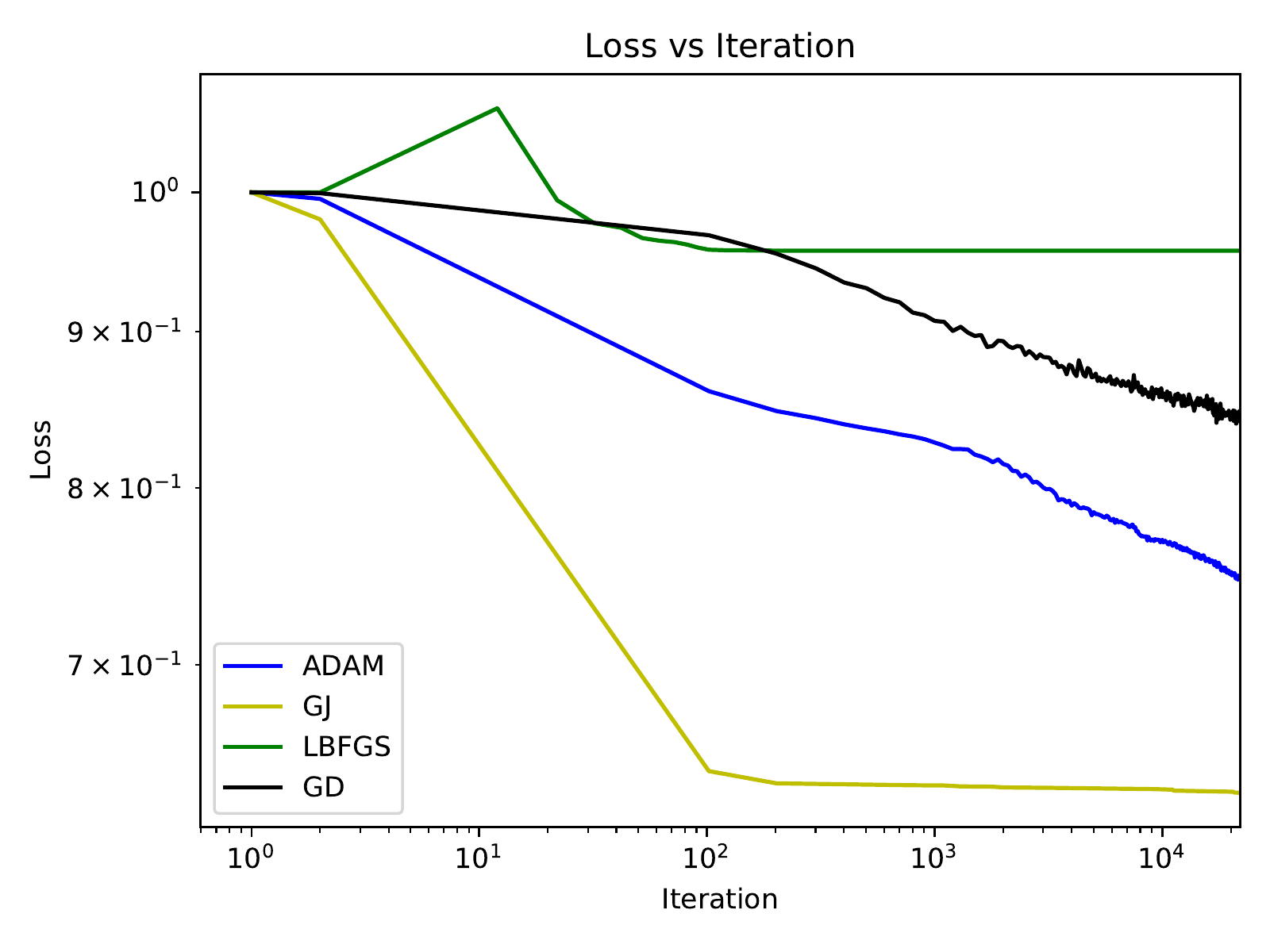}
        \\                 
        \includegraphics[width=.6\mylength]{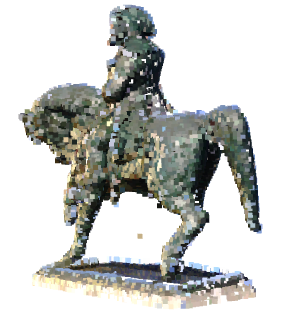} 
        &\includegraphics[width=\mylength]{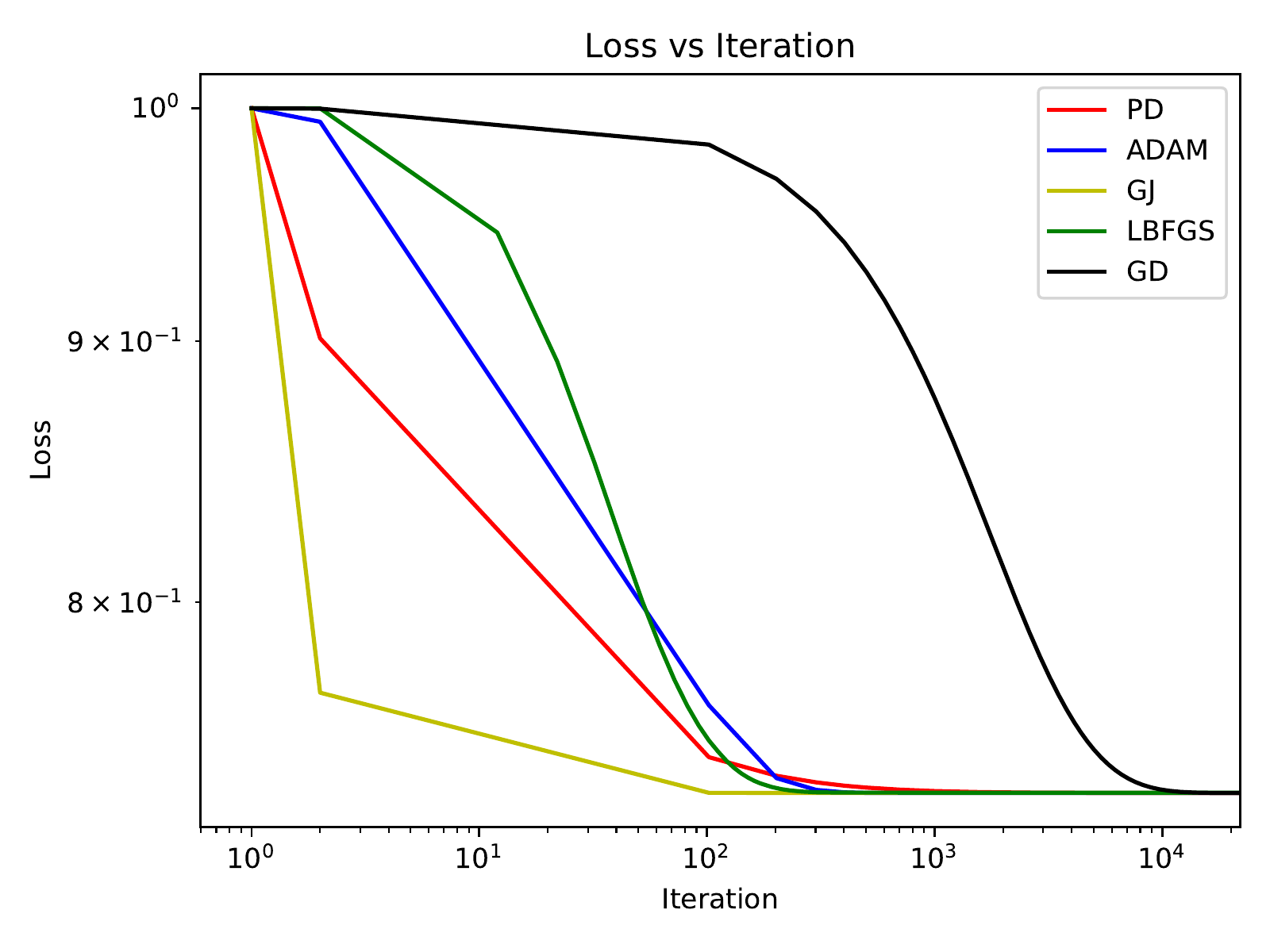}
        & \includegraphics[width=\mylength]{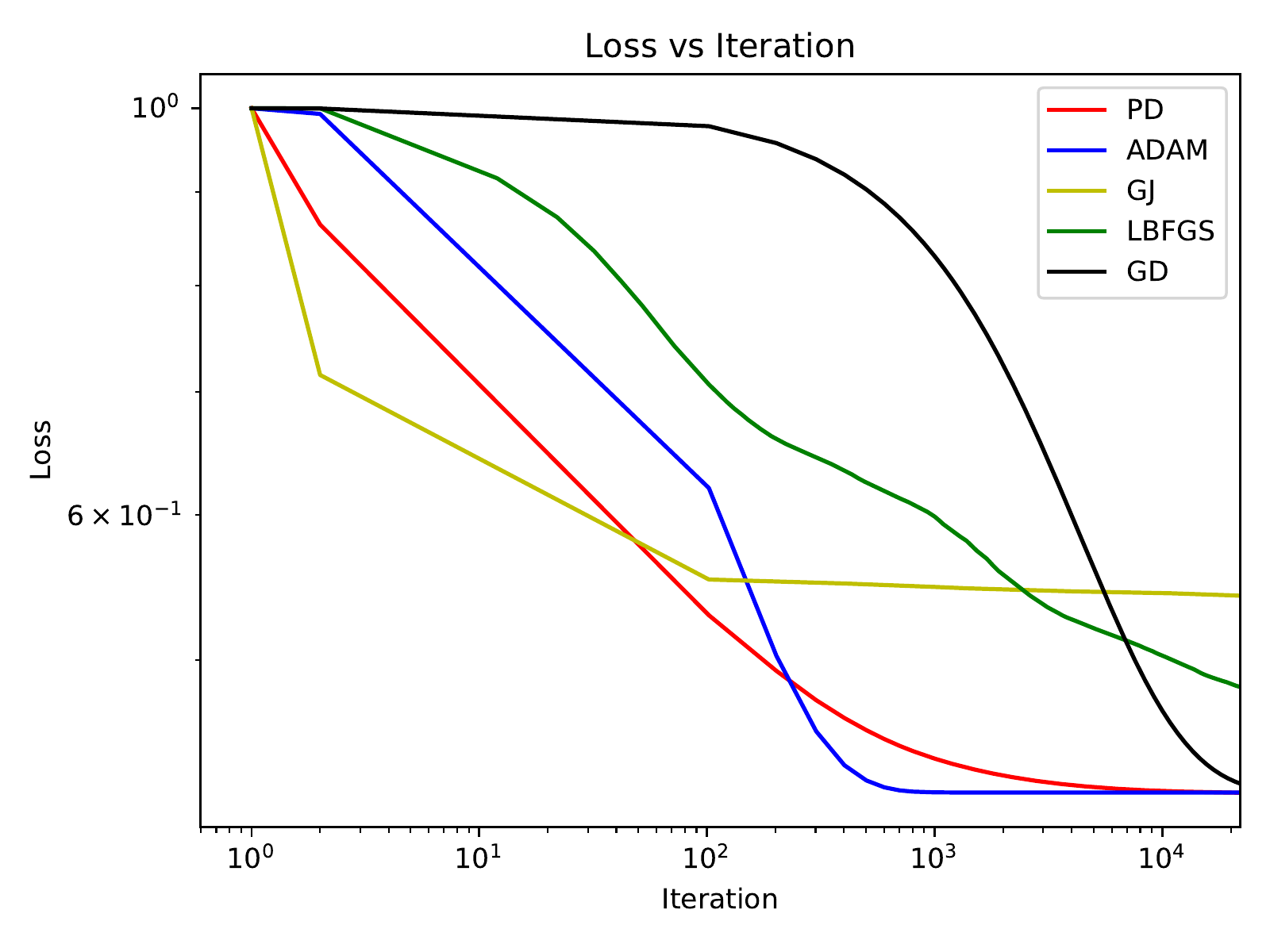}
        &
    \end{tabular}
    \caption{(left) Raw data $f_0$ used for experiments in Fig.~\ref{fig:elephant_inverse} and Fig.~\ref{fig:napoleon_inverse}. 
    (right) Comparison of the evolution of the objective function $J(f^{(t)})$ depending on iteration $(t)$, {in logarithmic scale}, for various setting of $(p,q)$ and for different algorithms. 
    }
    \label{fig:loss evolution}
\end{figure}

\setlength{\mylength}{0.17\linewidth}
\setlength{\tabcolsep}{3pt}
\renewcommand{\arraystretch}{1.}
\begin{figure}[htb]
    \centering
    \begin{tabular}{c cc ccc}
        & \textsc{mpn-pd} 
        & \textsc{mpn-gj} 
        & \textsc{gd} 
        & \textsc{adam} 
        & \textsc{lbfgs} 
        \\
        \raisebox{.35\mylength}{\rotatebox[origin=c]{90}{$q=1$}}
        \raisebox{.35\mylength}{\rotatebox[origin=c]{90}{$p=2$}}
        & \includegraphics[width=\mylength]{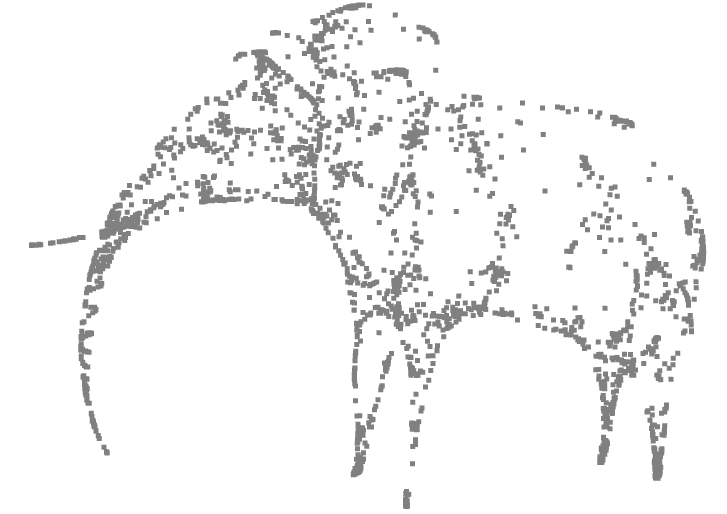}
        & \includegraphics[width=\mylength]{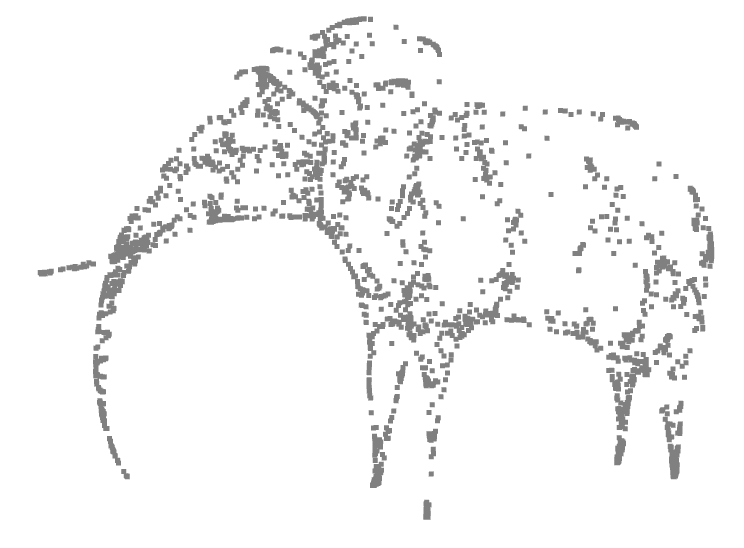}
        & \includegraphics[width=\mylength]{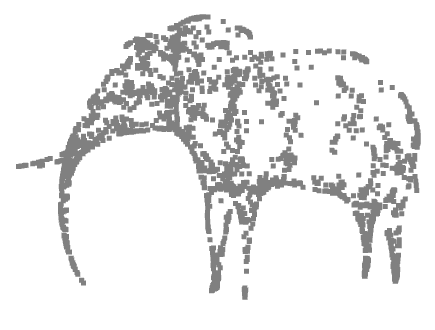}
        & \includegraphics[width=\mylength]{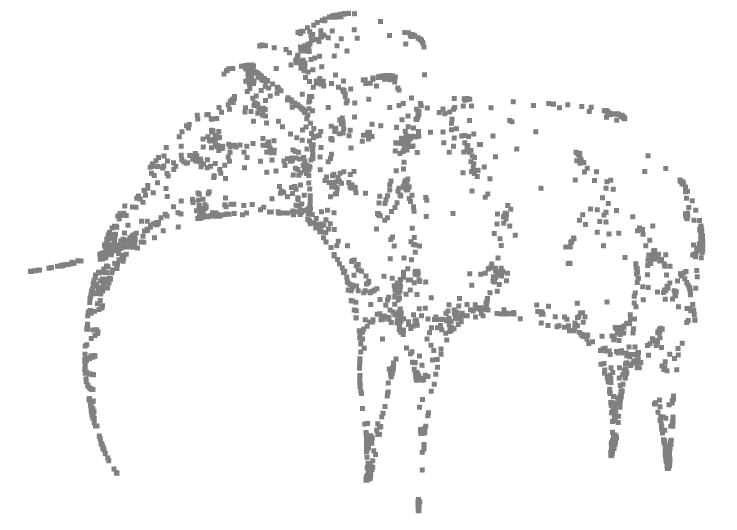}
        & \includegraphics[width=\mylength]{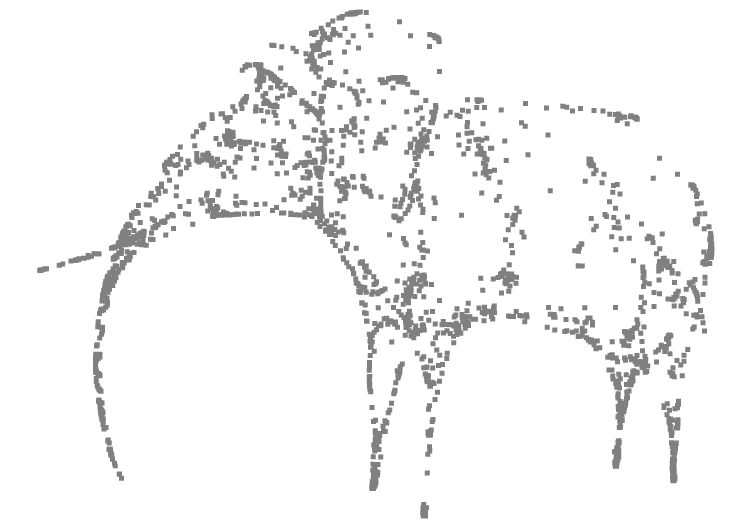}
        \\
        \raisebox{.35\mylength}{\rotatebox[origin=c]{90}{$q=1$}}
        \raisebox{.35\mylength}{\rotatebox[origin=c]{90}{$p=1$}}
        & \includegraphics[width=\mylength]{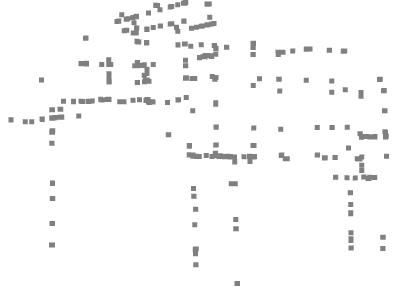}
        & \includegraphics[width=\mylength]{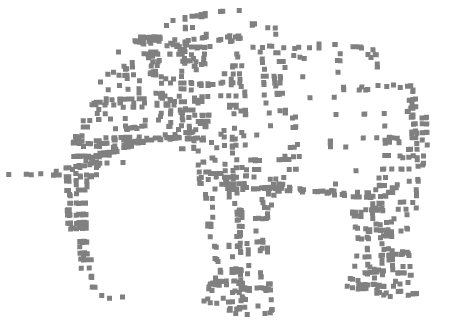}
        & \includegraphics[width=\mylength]{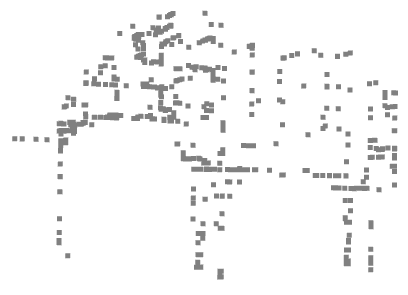} 
        & \includegraphics[width=\mylength]{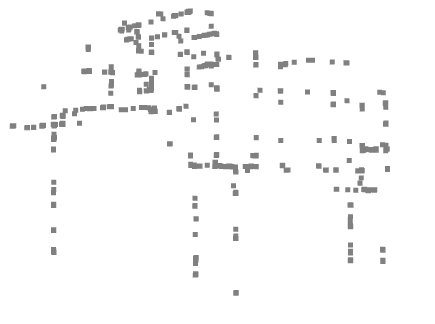}
        & \includegraphics[width=\mylength]{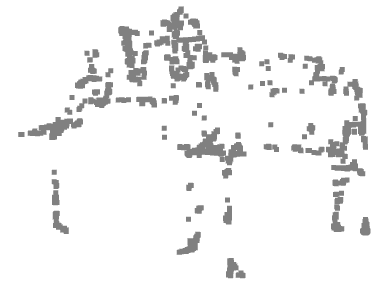}
        \\
        \raisebox{.35\mylength}{\rotatebox[origin=c]{90}{$q=0.1$}}
        \raisebox{.35\mylength}{\rotatebox[origin=c]{90}{$p=2$}}
        & 
        & \includegraphics[width=\mylength]{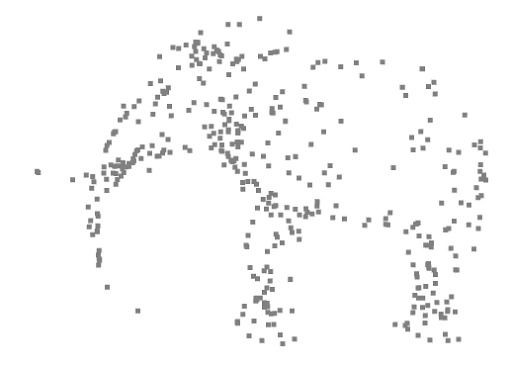}
        & \includegraphics[width=\mylength]{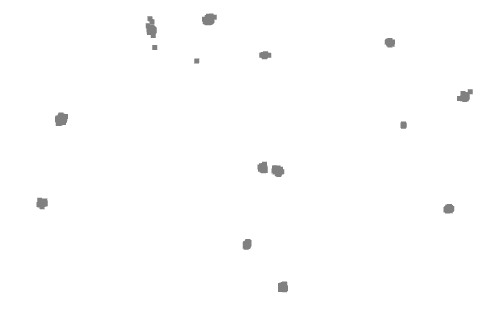} 
        & \includegraphics[width=\mylength]{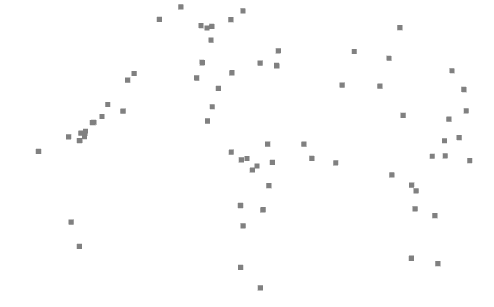}
        & \includegraphics[width=\mylength]{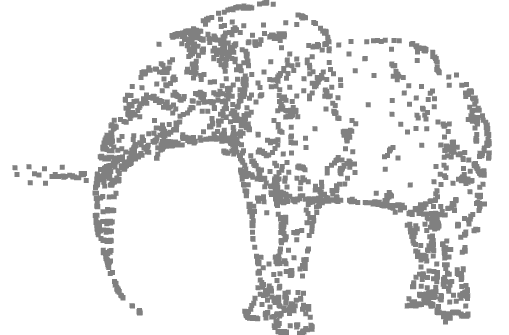}
        \\
    \end{tabular}
    \caption{
    Comparison of numerical results of MPN implementations of algorithms~\ref{algo:GJ} (\textsc{mpn-gj}) and~\ref{algo:CP} (\textsc{mpn-pd}), with \textsc{gd}, \textsc{adam} and \textsc{lbfgs} after $20k$ iterations. 
    Point-cloud simplification is defined by problem \eqref{eq:optimization_problem} with
    $\lambda=0.2$ and various settings of $p$ and $q$. 
    For all methods except \textsc{mpn-pd}, the approximated formulation \eqref{eq:NL-TV_eps} with $\epsilon=10^{-8}$ is used.
    }
    \label{fig:elephant_inverse}
\end{figure}

\setlength{\mylength}{0.17\linewidth}
\setlength{\tabcolsep}{3pt}
\renewcommand{\arraystretch}{1.}
\begin{figure}[!htb]
    \centering
    \begin{tabular}{c cc ccc}
        & \textsc{mpn-pd} 
        & \textsc{mpn-gj} 
        & \textsc{gd} 
        & \textsc{adam} 
        & \textsc{lbfgs} 
        \\
        \raisebox{.35\mylength}{\rotatebox[origin=c]{90}{$p=2$}} 
        & \includegraphics[width=\mylength]{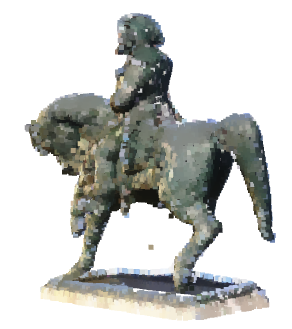}
        & \includegraphics[width=\mylength]{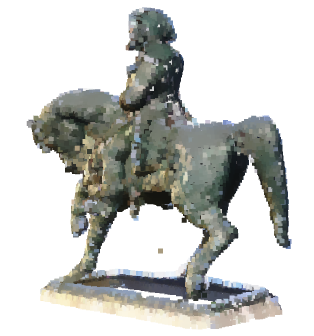}
        & \includegraphics[width=\mylength]{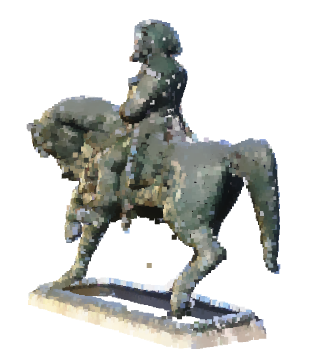}
        & \includegraphics[width=\mylength]{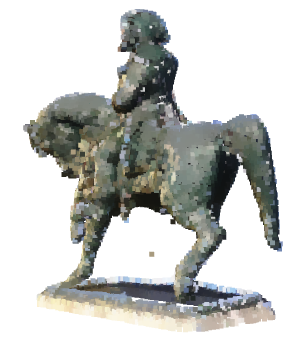}
        & \includegraphics[width=\mylength]{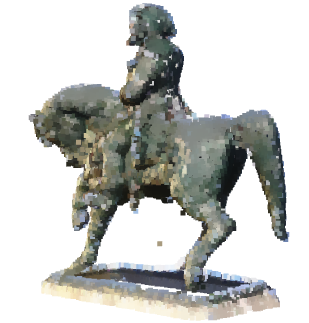}
        \\
        \raisebox{.35\mylength}{\rotatebox[origin=c]{90}{$p=1$}}
        & \includegraphics[width=\mylength]{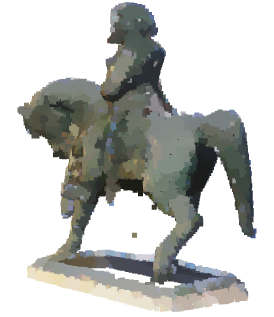}
        & \includegraphics[width=\mylength]{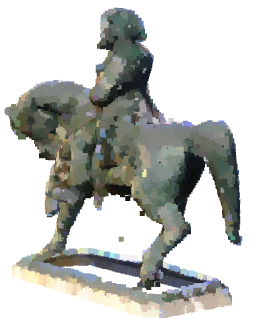}
        & \includegraphics[width=\mylength]{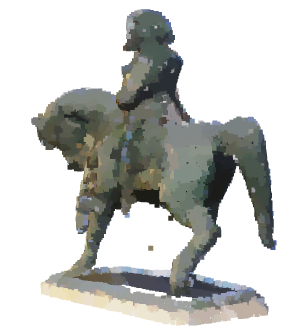} 
        & \includegraphics[width=\mylength]{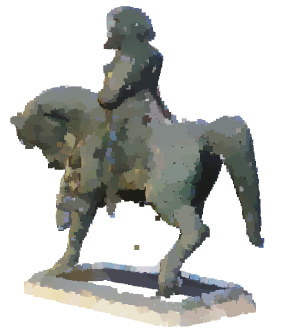}
        & \includegraphics[width=\mylength]{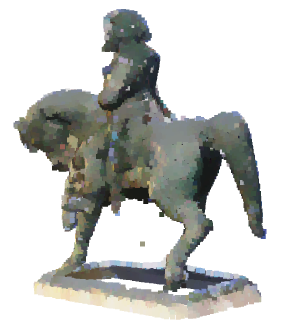}
    \end{tabular}
    \caption{Same setting as in Fig.~\ref{fig:elephant_inverse} except $\lambda = 0.05$, for color processing.
    }
    \label{fig:napoleon_inverse}
\end{figure}

\paragraph{Results.}
Observe that the noise on the raw color point cloud in Fig.~\ref{fig:loss evolution} results from the registration of multiple scans with different lighting conditions, for which we do not have any ground-truth.
In such a case, variational models like in \eqref{eq:optimization_problem} are useful to remove color artifacts without more specific prior knowledge.

Figures~\ref{fig:elephant_inverse} and~\ref{fig:napoleon_inverse} show the visual results of the aforementioned algorithms after 20k iterations. The objective function $J(f^{(t)})$ is plotted for each one in Fig.~\ref{fig:loss evolution}.
Since the computation time \emph{per} iteration is roughly same for all methods (about 0.5 ms for \textsc{mpn-pd} and 1 ms for others on `elephant' and `Napoléon' pointclouds with $|\V|= 3k$ and $11k$ respectively), curves are here simply displayed versus the number of iterations.
As one can observe from these experiments, MPN based algorithms perform better than gradient based algorithm when it comes to precision. Indeed, for $q=1$, Alg.~\ref{algo:CP} is the only algorithm that can be used without approximation (\ie $\varepsilon=0$, which is numerically intractable for other methods) and that gives ultimately an optimal solution. 
As for $q=0.1$, Alg.~\ref{algo:GJ} reaches an (approximated) solution much faster than gradient based methods. 
Among these, it is interesting to notice that \textsc{adam} is quite consistent and performs quite well. For this reason, we will consider this method for the next part devoted to unsupervised training of GNN.

\smallskip
We have shown that the two algorithms can be implemented efficiently as MPN, a special instance of GNN.
In the next section, we demonstrate the interest of training such MPNs to design feed-forward GNNs allowing fast processing. 
\section{Variational training of GNN}
\label{sec:GNN}

One of the most attractive aspect of machine learning is the ability to learn the model itself from the data. As already mentioned in the introduction, most approaches rely on \emph{supervised} training techniques which require datasets with ground-truth, potentially obtained by means of data augmentation (\ie artificially degrading the data), which might be a simple task for some applications such as Gaussian denoising on images. 
However, it is challenging and time consuming to achieve this for signal processing tasks on point-clouds. 
Besides, most of the available datasets are only devoted to high-level tasks like segmentation and classification on shapes.

Inspired from the previous results, we investigate the \emph{unsupervised} training of a GNN to reproduce the behavior of variational models on point-clouds while reducing the computation time.
To achieve such a goal, two different settings are considered in this section:
Unsupervised training using the variational energy as a loss function (\S~\ref{sec:unsupervised}),
and model distillation using exemplars from the variational model defined as an MPN (\S~\ref{sec:distillation}).

\subsection{Unsupervised training}\label{sec:unsupervised}
We denote by $G_\theta : \R^{|\V|\times d} \times \R_+^{\E} \to \R^{|\V|\times d}$ a trainable GNN parametrized by $\theta$,
processing a graph defined by the vertex features $f$ and the edge weights $\omega$. 
The idea here is simply to train a GNN in such a way that it minimizes in expectation the objective function \eqref{eq:optimization_problem} of the variational model for graphs from a dataset
\begin{equation}\label{eq:unsupervised_training}
    \inf_\theta 
        \mathbb{E}_{X \sim P_\text{data}}
        J \left( G_\theta(X,\omega(X)) \right).
\end{equation}
where $X$ is a random point-cloud (and its associated weights $\omega(X)$) from the training dataset with probability distribution $P_\text{data}$.
Once again, in practice, we need to consider the $\varepsilon$-approximation $J_\varepsilon$ \eqref{eq:NL-TV_eps} to overcome numerical issues.

\subsection{Model distillation with MPN}\label{sec:distillation}
As seen above, the main caveat in variational training is that it requires smoothing if the objective function is not differentiable. 
to avoid this limitation, we consider another setting which consists of model distillation, where the goal is to train a gnn to reproduce an exact model defined by a mpn.

Let $F : \R^{|\V|\times d} \times \R_+^{\E} \mapsto \R^{|\V|\times d}$ denotes a MPN designed to solve a variational model, such as \textsc{mpn-pd} in the previous section for the NL-TV model.
Note that in practice, we have to restrict this network to $n$ updates. 
Using for instance $\ell_2$ norm, the distillation training boils down to the following optimization problem
\begin{equation}\label{eq:distillation_training}
    \inf_\theta 
        \mathbb{E}_{X \sim P_\text{data}}
        \lVert G_\theta(X,\omega(X)) - F(X,\omega(X))\rVert^2
\end{equation}
Other choices of distance could be as well considered, such as $\ell_1$ norm or optimal transport cost for instance \cite{williams2019deep}. 

\subsection{Experiments}
\paragraph{Dataset setting.} 
We use the same experimental setting as in the previous section, except that $k$ nearest-neighbor parameters $\omega_{i,j} \in \{0,1\}$ are now weighted using features proximity: $\omega_{i,j} \leftarrow \omega_{i,j} e^{- \kappa \|(f_0)_i - (f_0)_j\|^2}$ where $\kappa = 10^{4}$.
For point-cloud processing, we use the ShapeNet part dataset~\cite{shapenetpart}.
Point-clouds are pre-processed by normalizing {their features to 0 and 1 using minmax normalization};
15 classes out of 16 are used for training and a hold out class for testing. For each training class, 20\% of shapes are also hold-out for validation. 
In total, 16.5k point clouds are used during training, composed of an average of 2.5k nodes.

\paragraph{GNN architecture.}
The GNN is based on  MPNs. The architecture is shallow and composed of 3 MPN layers followed by a linear layer. 
Inspired from MPNs formulated in Section~\ref{MPN} from Alg~.\ref{algo:GJ} and Alg~.\ref{algo:CP}, we define the message function ($\phi$) as a MLP acting on the tuple $(f_{i}, f_{j}, \sqrt{w_{i,j}}(f_{j} - f_{i}))$.
Note that, this is quite similar to edge-convolution operation introduced in~\cite{wang2019dynamic}, in which the MLP layers act on features $(f_{i}, f_{j} - f_{i}, w_{i,j})$.
The $\square$ operator is $\sum$ and there is no update rule ($\psi$). 

The MLP layers have respectively 4.8k, 20.6k, and 20.6k parameters. 
These layers operate on 64 dimensional features, which are then fed to a linear layer to return 3-dimensional features (color or coordinates). 

\paragraph{Training settings.}
Using the dataset and the architecture described above, two GNNs are trained using each of the techniques described in \S~\ref{sec:unsupervised} and \S~\ref{sec:distillation}, with $k=4$, $p=2$, $q=1$, $\lambda = 0.05$ and $\varepsilon=10^{-8}$.
Training is performed using \textsc{Adam} algorithm with 60 epochs and a fixed learning rate of $0.01$. 
Regarding the model distillation, rather than re-evaluating the MPN $F$ in \eqref{eq:distillation_training} for every sample $X$ at each epoch, we precompute $F(X,\omega(X))$ to save the computation time. 

\paragraph{Results.} Fig.~\ref{fig:GNN point cloud} gives a visual comparison of results of trained GNNs with the exact \textsc{mpn-pd} optimization on test data, showing that both training methods give satisfactory results, close to \textsc{mpn-pd} model.
Fig.~\ref{fig:GNN evolution} displays
the objective loss function in \eqref{eq:unsupervised_training} during training, on the training and the validation set. 
The average error relative to \text{mpn-pd} on the test set is similar for both methods ($2.2\%$ for distillation and $1.2\%$ for unsupervised training).
One can observe that model distillation is more stable on the validation set, but need extra computation time to pre-process the \textsc{mpn-pd} on the dataset.
Last, but not least, the average computation time for \textsc{mpn-pd} on the test set is compared in Fig.~\ref{fig:GNN evolution} to the feed-forward GNNs, showing several order of magnitude speed-up for the latter, which makes such approaches very interesting for practical use.

\setlength{\mylength}{0.24\linewidth}
\setlength{\tabcolsep}{1pt}
\renewcommand{\arraystretch}{1.}
\begin{figure}[htb]
    \centering
    \begin{tabular}{c c c c c}     
        &\multirow{2}{*}{Input $X$}
        & \textsc{mpn-pd} 
        & \textsc{gnn}
        & \textsc{gnn}
        \\ 
        &
        & (Alg.~\ref{algo:CP})
        &(\textsc{pd} distillation)
        &(Unsupervised)
        \\
        \raisebox{.35\mylength}{\rotatebox[origin=c]{90}{unseen}}
        &\includegraphics[width=0.8\mylength]{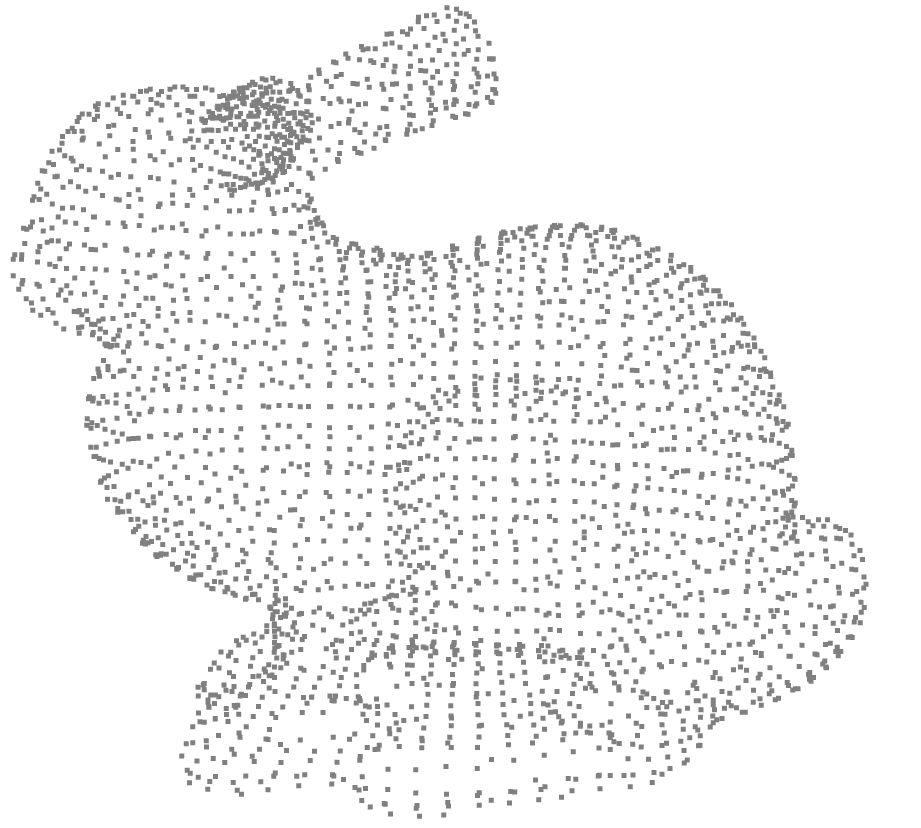}
        &\includegraphics[width=0.8\mylength]{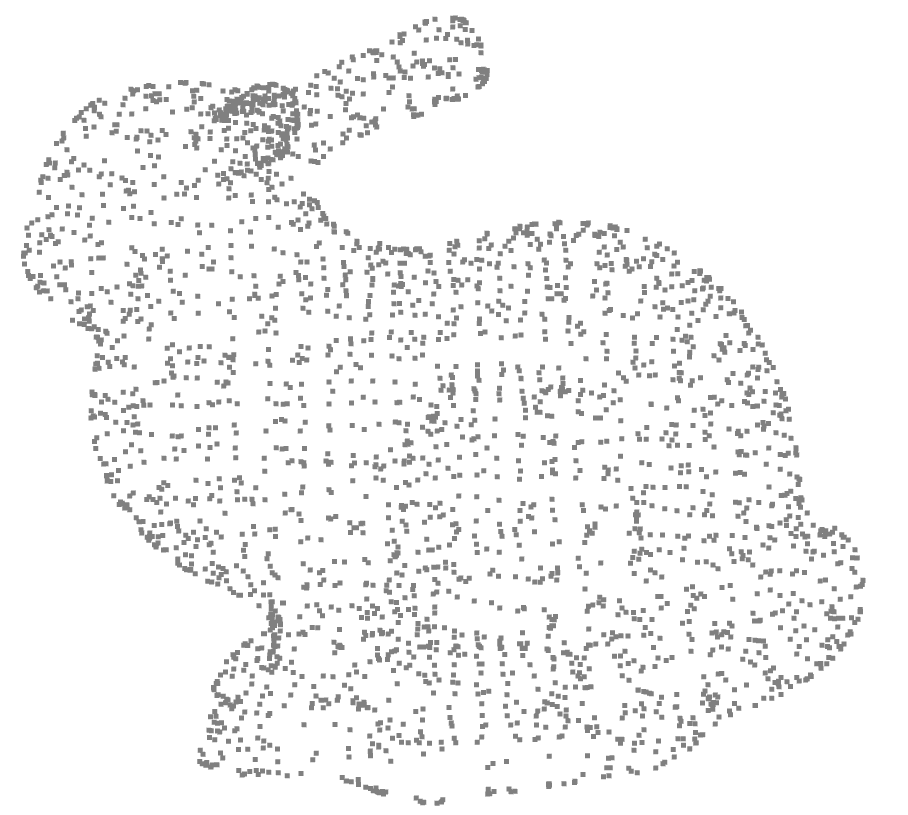}
        & \includegraphics[width=0.8\mylength]{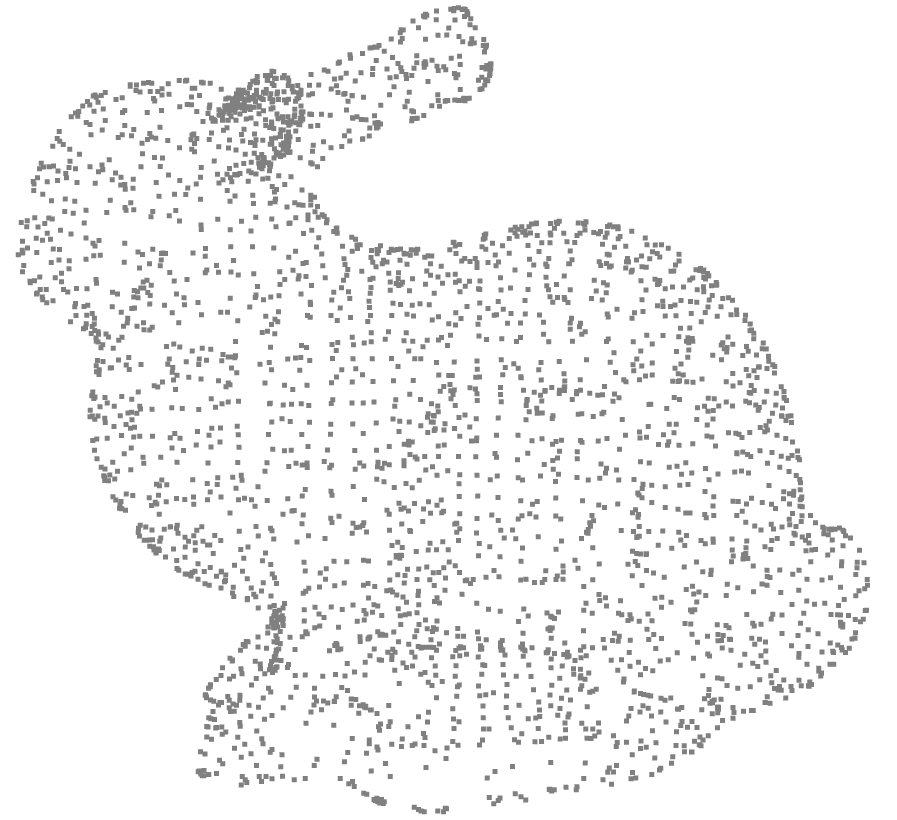}
        & \includegraphics[width=0.8\mylength]{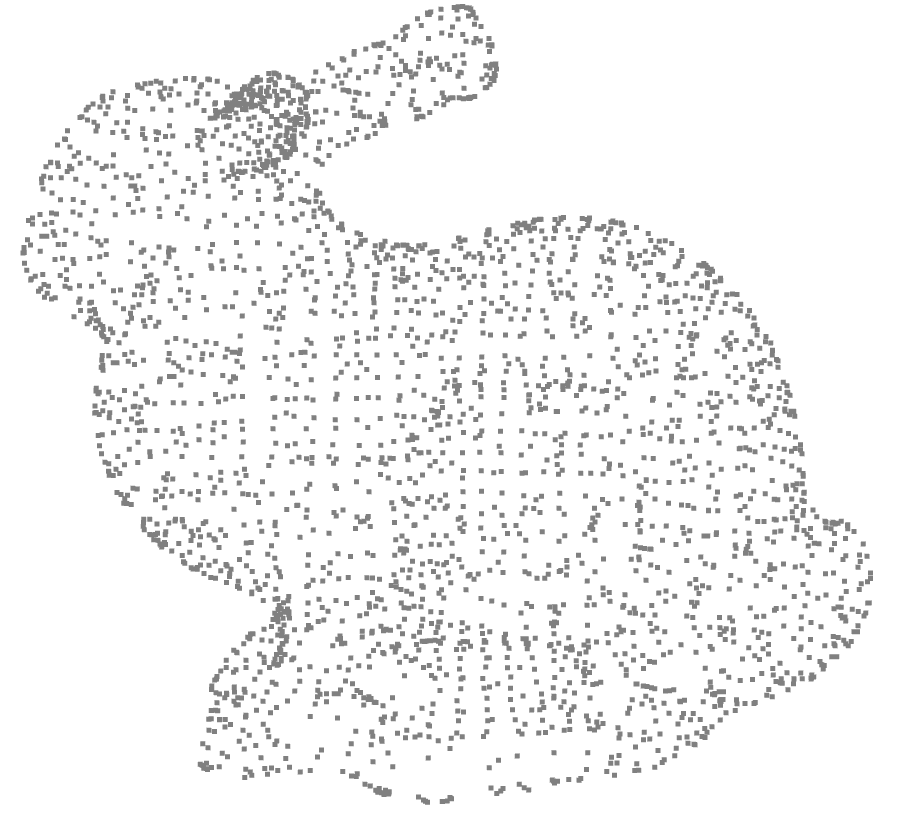}
        \\    
        \raisebox{.35\mylength}{\rotatebox[origin=c]{90}{unseen}}        
        &\includegraphics[width=\mylength]{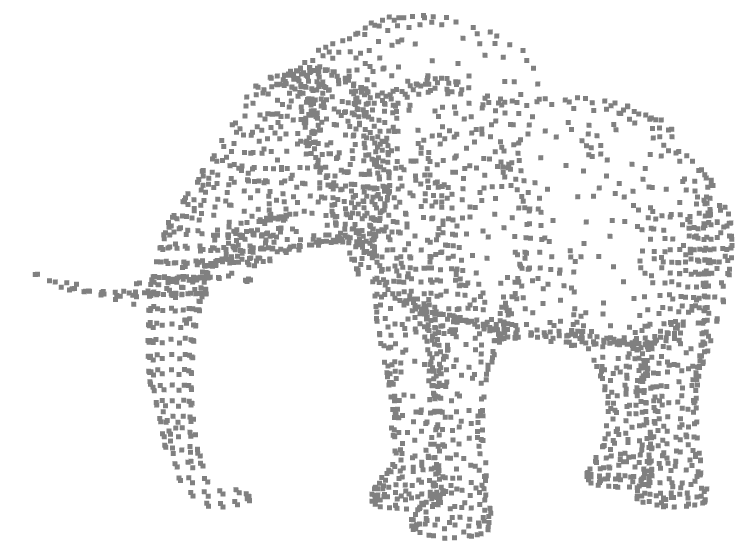}
        &\includegraphics[width=\mylength]{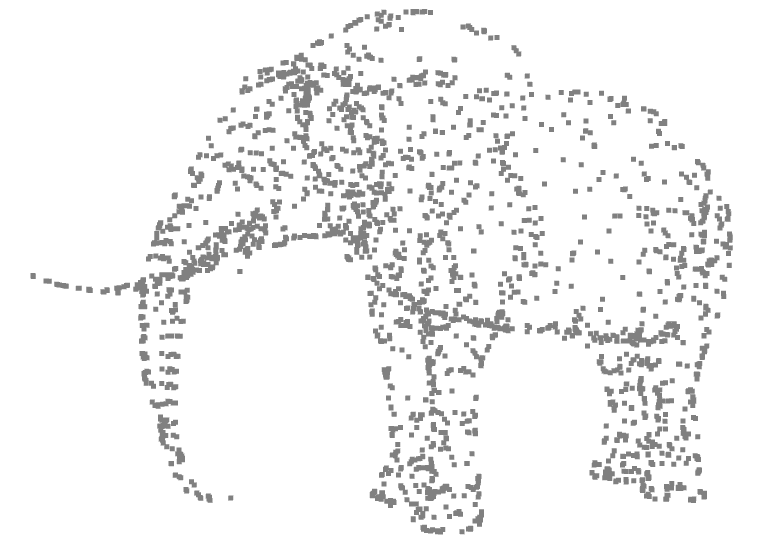}
        & \includegraphics[width=\mylength]{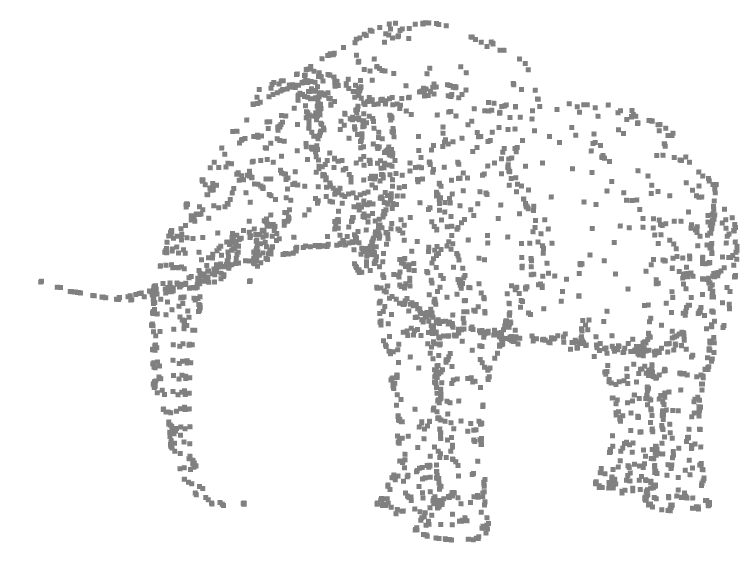}
        & \includegraphics[width=\mylength]{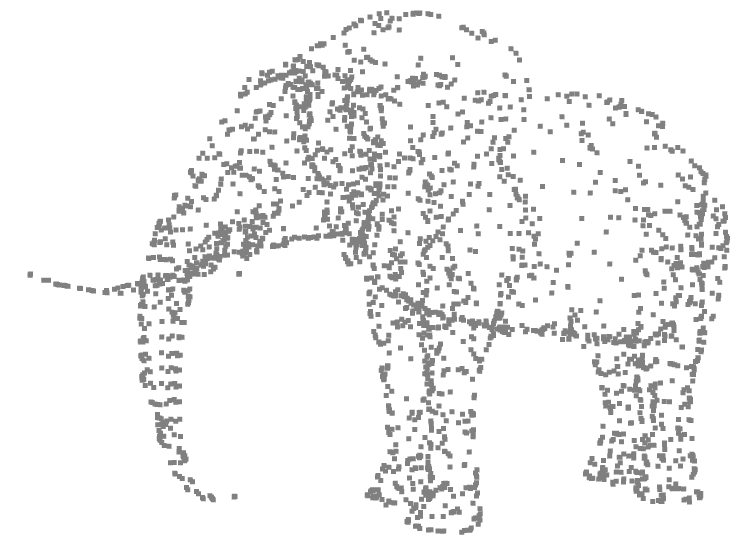}
        \\     
        \raisebox{.2\mylength}{\rotatebox[origin=c]{90}{test}}
        &\includegraphics[width=\mylength]{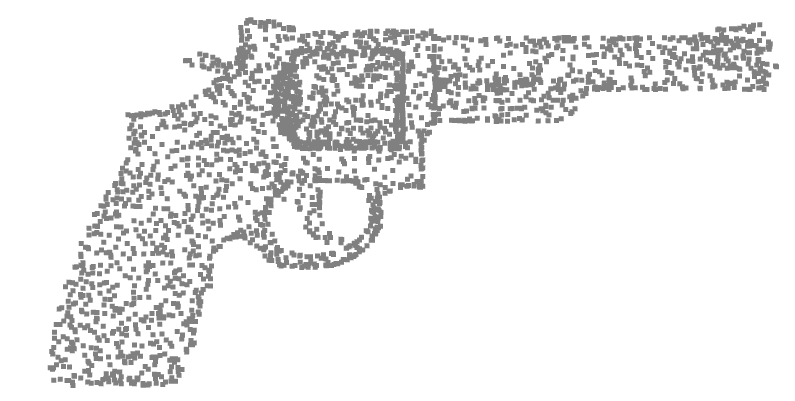}
        &\includegraphics[width=\mylength]{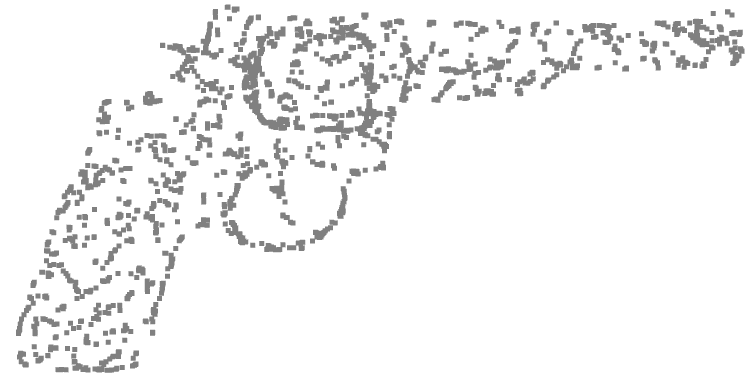}
        & \includegraphics[width=\mylength]{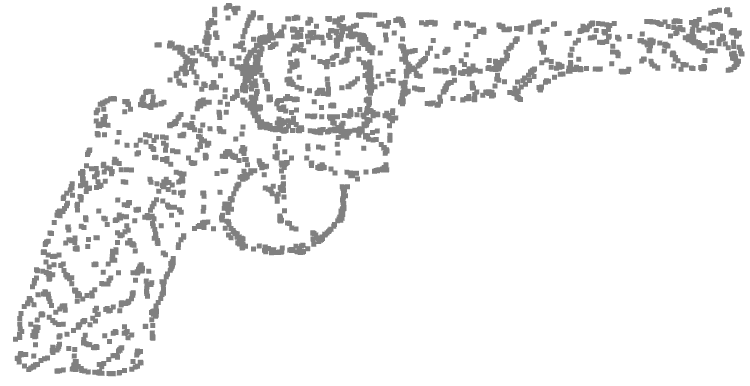}
        & \includegraphics[width=\mylength]{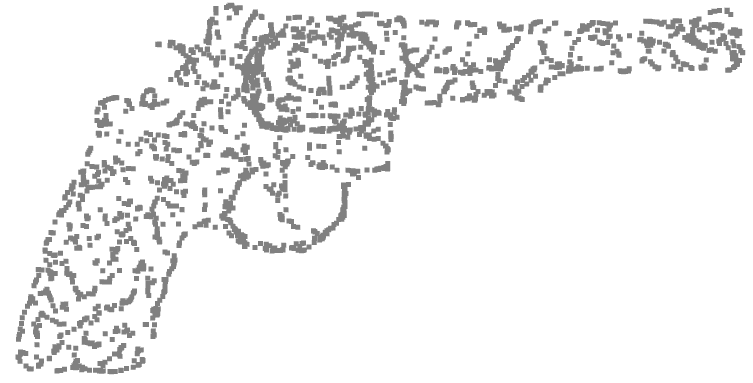}
        
    \end{tabular}
    \caption{Comparison of NL-TV regularization \eqref{eq:optimization_problem} of point-clouds $X$ (for $p=2$, $q=1$, $\lambda = 0.05$) with the MPN algorithm \textsc{mpn-pd} and trained GNN, using model distillation \eqref{eq:distillation_training} or the unsupervised setting with variational optimization \eqref{eq:unsupervised_training}.
    Input here is either \emph{unseen} point-clouds during training, or from the \emph{test} category, hold out during training.
    }
    \label{fig:GNN point cloud}
\end{figure}
\setlength{\mylength}{0.32\linewidth}
\setlength{\tabcolsep}{3pt}
\renewcommand{\arraystretch}{1.}
\begin{figure}[!htb]
    \centering
    \begin{tabular}{c c  | c}
        Training set
        & Validation set & Test set
        \\
        \includegraphics[width=\mylength]{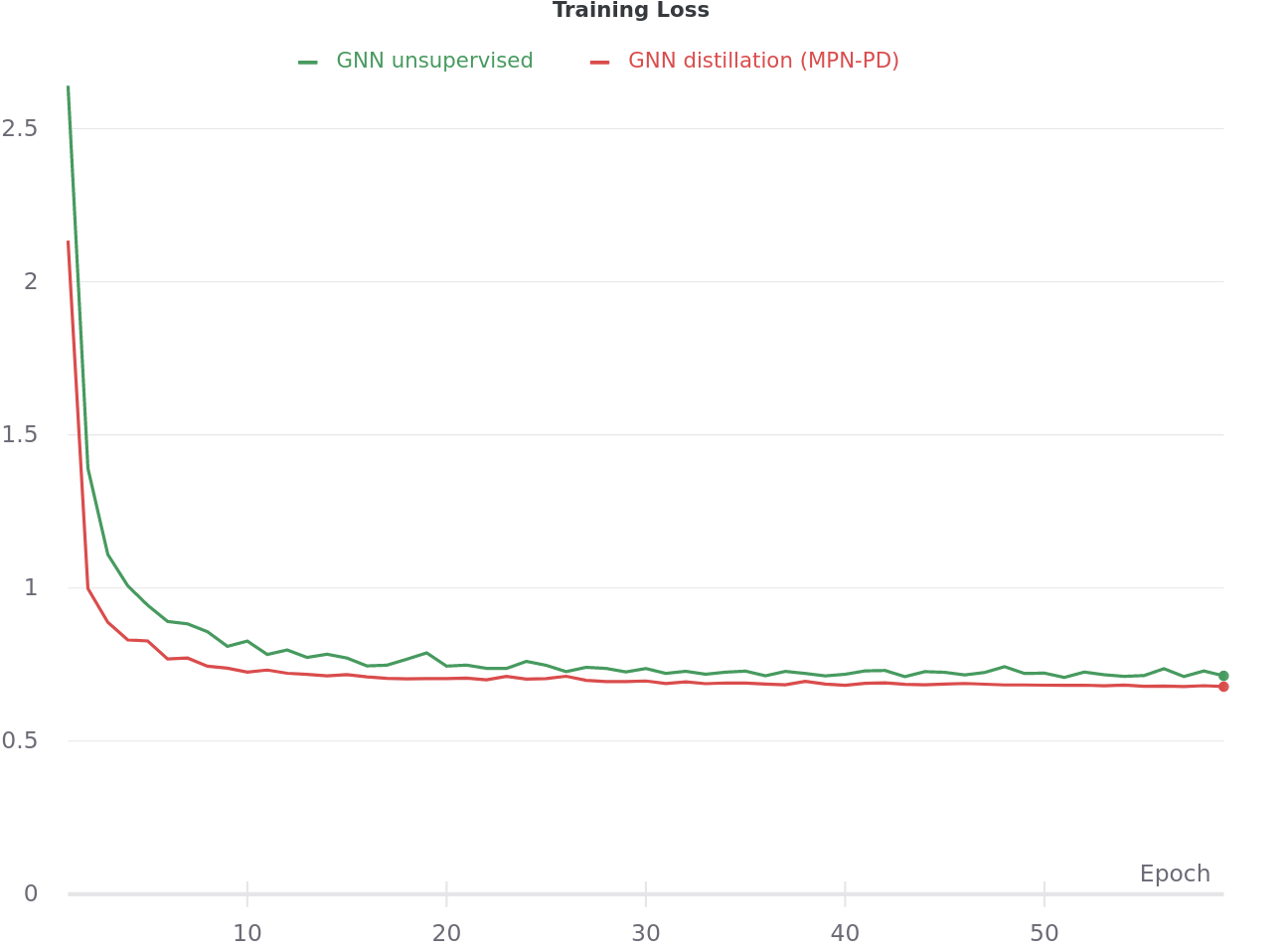}
        &\includegraphics[width=\mylength]{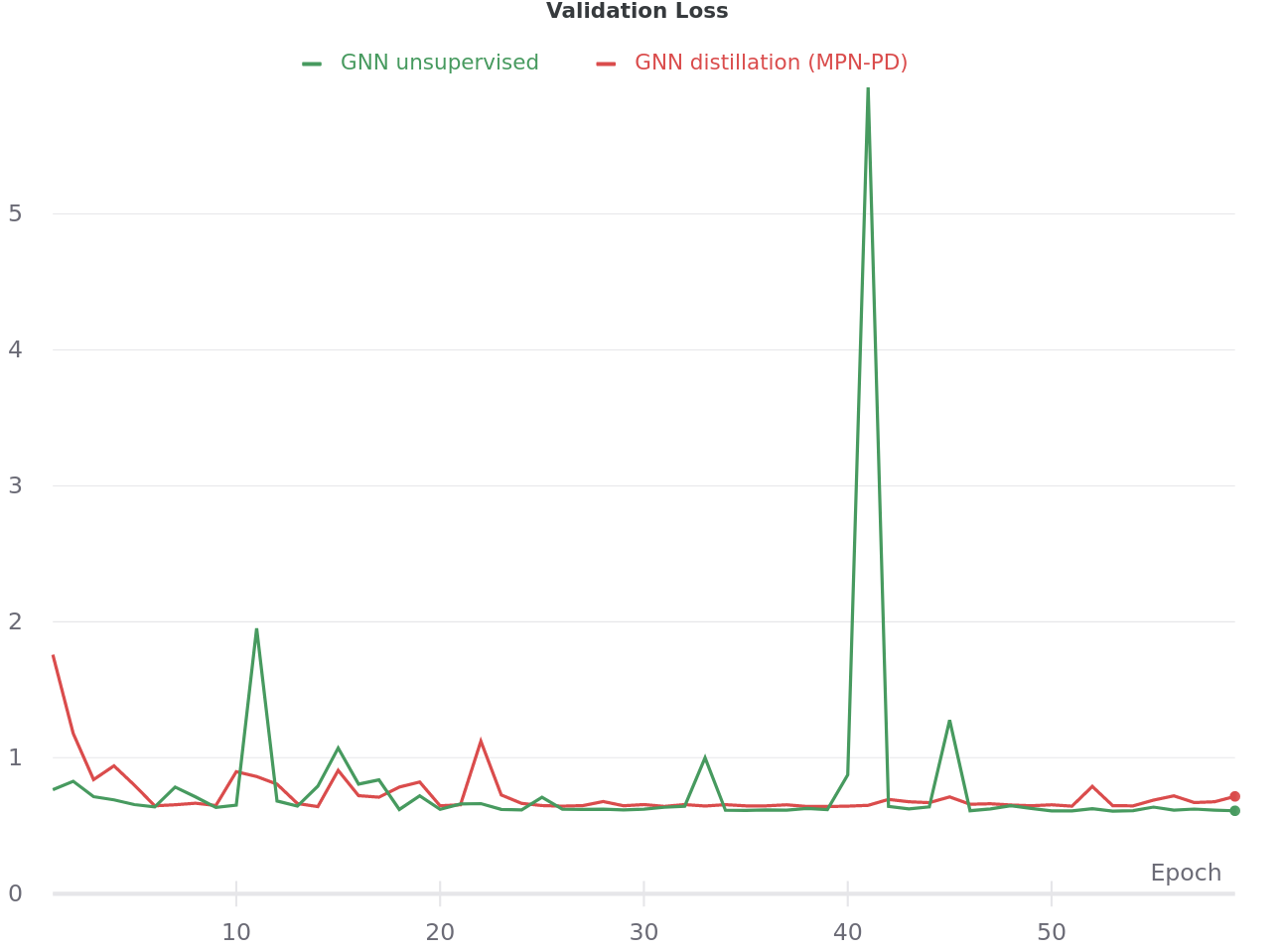}
        &\includegraphics[width=\mylength]{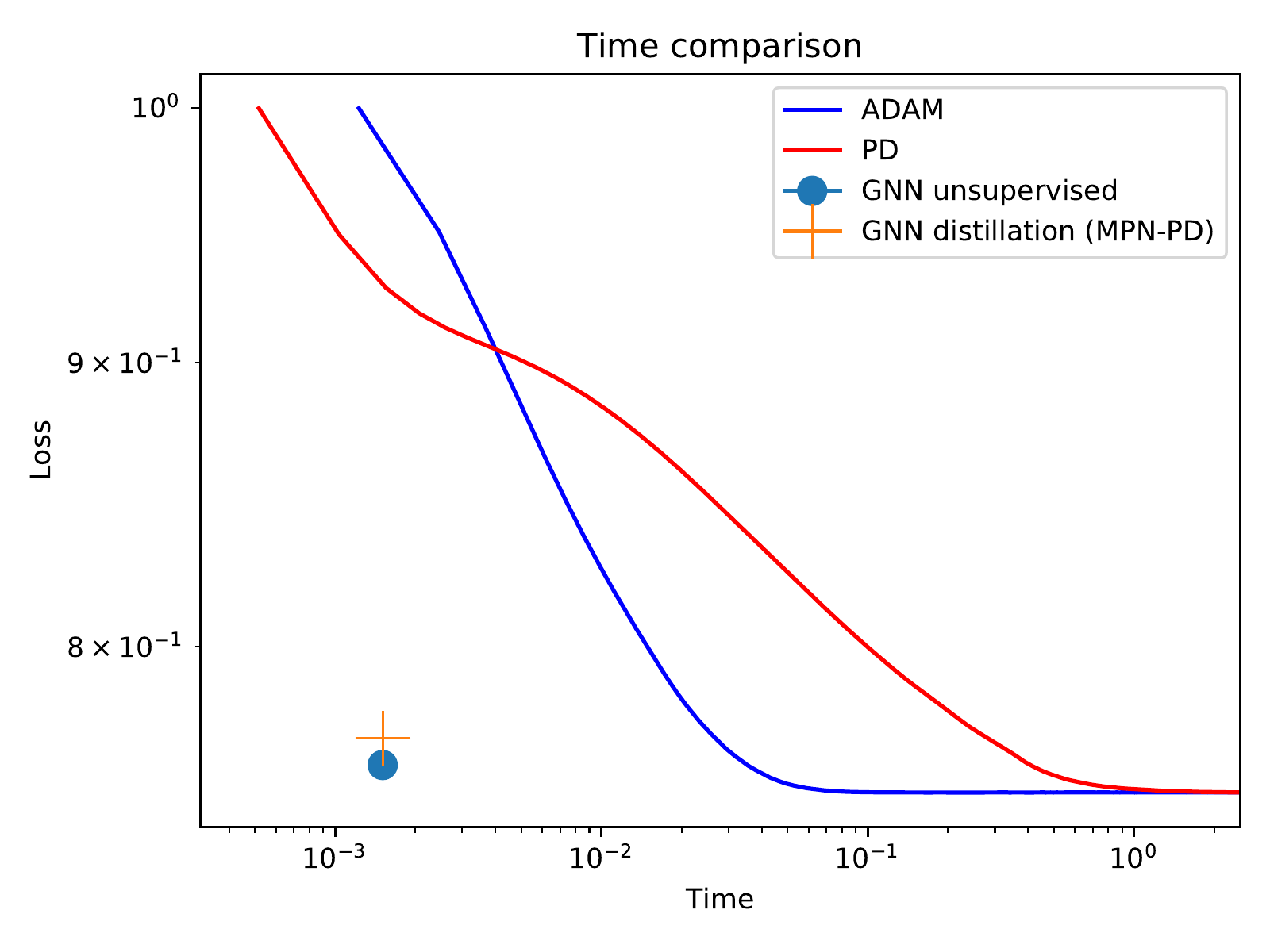}
    \end{tabular}
    \caption{(left) Objective loss function \eqref{eq:unsupervised_training} during GNN training versus epochs for the training and the validation sets.
    (right) Comparison of the average objective function $J$ {in logarithmic scale} versus \emph{average computation time} for GNN, \textsc{adam} and \textsc{mpn-pd}.
    }
    \label{fig:GNN evolution}
\end{figure}

\noindent\textbf{Acknowledgments} This work has been carried out with financial support from the French Research Agency through the SUMUM project (ANR-17-CE38-0004)


\bibliographystyle{splncs04}
\bibliography{bib}

\begin{thebibliography}{10}
\providecommand{\url}[1]{\texttt{#1}}
\providecommand{\urlprefix}{URL }
\providecommand{\doi}[1]{https://doi.org/#1}

\bibitem{bertocchi2020deep}
Bertocchi, C., Chouzenoux, E., Corbineau, M.C., Pesquet, J.C., Prato, M.: Deep
  unfolding of a proximal interior point method for image restoration. Inverse
  Problems  \textbf{36}(3),  034005 (2020)

\bibitem{bruna2014spectral}
Bruna, J., Zaremba, W., Szlam, A., LeCun, Y.: Spectral networks and locally
  connected networks on graphs. In: Int. Conf. on Learning Representations
  (2014)

\bibitem{chambolle2011first}
Chambolle, A., Pock, T.: A first-order primal-dual algorithm for convex
  problems with applications to imaging. Journal of mathematical imaging and
  vision  \textbf{40}(1),  120--145 (2011)

\bibitem{combettes2020deep}
Combettes, P.L., Pesquet, J.C.: Deep neural network structures solving
  variational inequalities. Set-Valued and Variational Analysis pp. 1--28
  (2020)

\bibitem{defferrard2016convolutional}
Defferrard, M., Bresson, X., Vandergheynst, P.: Convolutional neural networks
  on graphs with fast localized spectral filtering. In: Advances in neural
  information processing systems. pp. 3844--3852 (2016)

\bibitem{elmoataz2008nonlocal}
Elmoataz, A., Lezoray, O., Bougleux, S.: Nonlocal discrete regularization on
  weighted graphs: a framework for image and manifold processing. IEEE
  Transactions on Image Processing  \textbf{17}(7),  1047--1060 (2008)

\bibitem{elmoataz2015p}
Elmoataz, A., Toutain, M., Tenbrinck, D.: On the $p$-laplacian and
  $\infty$-laplacian on graphs with applications in image and data processing.
  SIAM Journal on Imaging Sciences  \textbf{8}(4),  2412--2451 (2015)

\bibitem{torchgeo}
Fey, M., Lenssen, J.E.: Fast graph representation learning with pytorch
  geometric. arXiv preprint arXiv:1903.02428  (2019)

\bibitem{gilboa2009nonlocal}
Gilboa, G., Osher, S.: Nonlocal operators with applications to image
  processing. Multiscale Modeling \& Simulation  \textbf{7}(3),  1005--1028
  (2009)

\bibitem{gilmer2017neural}
Gilmer, J., Schoenholz, S.S., Riley, P.F., Vinyals, O., Dahl, G.E.: Neural
  message passing for quantum chemistry. In: ICML (2017)

\bibitem{hasannasab2020parseval}
Hasannasab, M., Hertrich, J., Neumayer, S., Plonka, G., Setzer, S., Steidl, G.:
  Parseval proximal neural networks. Journal of Fourier Analysis and
  Applications  \textbf{26}(4),  1--31 (2020)

\bibitem{hidane2013nonlinear}
Hidane, M., L{\'e}zoray, O., Elmoataz, A.: Nonlinear multilayered
  representation of graph-signals. Journal of mathematical imaging and vision
  \textbf{45}(2),  114--137 (2013)

\bibitem{kingma2014adam}
Kingma, D.P., Ba, J.: Adam: A method for stochastic optimization. In:
  Proceedings of the 3rd International Conference on Learning Representations
  (ICLR) (2014)

\bibitem{kipf2017semi}
Kipf, T.N., Welling, M.: Semi-supervised classification with graph
  convolutional networks. In: Int. Conf. on Learning Representations (2017)

\bibitem{kobler2017variational}
Kobler, E., Klatzer, T., Hammernik, K., Pock, T.: Variational networks:
  connecting variational methods and deep learning. In: German conference on
  pattern recognition. pp. 281--293. Springer (2017)

\bibitem{li2019pu}
Li, R., Li, X., Fu, C.W., Cohen-Or, D., Heng, P.A.: Pu-gan: a point cloud
  upsampling adversarial network. In: Proceedings of the IEEE/CVF International
  Conference on Computer Vision. pp. 7203--7212 (2019)

\bibitem{liu1989limited}
Liu, D.C., Nocedal, J.: On the limited memory bfgs method for large scale
  optimization. Mathematical programming  \textbf{45}(1),  503--528 (1989)

\bibitem{lozes2013nonlocal}
Lozes, F., Hidane, M., Elmoataz, A., L{\'e}zoray, O.: Nonlocal segmentation of
  point clouds with graphs. In: 2013 IEEE Global Conference on Signal and
  Information Processing. pp. 459--462. IEEE (2013)

\bibitem{meinhardt2017learning}
Meinhardt, T., Moller, M., Hazirbas, C., Cremers, D.: Learning proximal
  operators: Using denoising networks for regularizing inverse imaging
  problems. In: Proceedings of the IEEE ICCV. pp. 1781--1790 (2017)

\bibitem{peyre2011non}
Peyr{\'e}, G., Bougleux, S., Cohen, L.: Non-local regularization of inverse
  problems. Inverse Problems \& Imaging  \textbf{5}(2), ~511 (2011)

\bibitem{qi2016pointnet}
Qi, C.R., Su, H., Mo, K., Guibas, L.J.: Pointnet: Deep learning on point sets
  for 3d classification and segmentation. In: Proceedings of the IEEE
  conference on computer vision and pattern recognition. pp. 652--660 (2017)

\bibitem{qi2017pointnetplusplus}
Qi, C.R., Yi, L., Su, H., Guibas, L.J.: Pointnet++: Deep hierarchical feature
  learning on point sets in a metric space. In: Advances in Neural Information
  Processing Systems. vol.~30 (2017)

\bibitem{raguet2015preconditioning}
Raguet, H., Landrieu, L.: Preconditioning of a generalized forward-backward
  splitting and application to optimization on graphs. SIAM Journal on Imaging
  Sciences  \textbf{8}(4),  2706--2739 (2015)

\bibitem{raguet2018cut}
Raguet, H., Landrieu, L.: Cut-pursuit algorithm for regularizing nonsmooth
  functionals with graph total variation. In: International Conference on
  Machine Learning. pp. 4247--4256 (2018)

\bibitem{tabti2018symmetric}
Tabti, S., Rabin, J., Elmoataz, A.: Symmetric upwind scheme for discrete
  weighted total variation. In: IEEE International Conference on Acoustics,
  Speech and Signal Processing (ICASSP). pp. 1827--1831 (2018)

\bibitem{tenbrinck2019variational}
Tenbrinck, D., Gaede, F., Burger, M.: Variational graph methods for efficient
  point cloud sparsification. arXiv preprint arXiv:1903.02858  (2019)

\bibitem{wang2019deep}
Wang, Y., Solomon, J.M.: Deep closest point: Learning representations for point
  cloud registration. In: Proceedings of the IEEE International Conference on
  Computer Vision. pp. 3523--3532 (2019)

\bibitem{wang2019dynamic}
Wang, Y., Sun, Y., Liu, Z., Sarma, S.E., Bronstein, M.M., Solomon, J.M.:
  Dynamic graph cnn for learning on point clouds. ACM Transactions On Graphics
  (TOG)  \textbf{38}(5),  1--12 (2019)

\bibitem{williams2019deep}
Williams, F., Schneider, T., Silva, C., Zorin, D., Bruna, J., Panozzo, D.: Deep
  geometric prior for surface reconstruction. In: Proceedings of the IEEE
  Conference on Computer Vision and Pattern Recognition. pp. 10130--10139
  (2019)

\bibitem{pmlr-v97-wu19e}
Wu, F., Souza, A., Zhang, T., Fifty, C., Yu, T., Weinberger, K.: Simplifying
  graph convolutional networks. In: Proceedings of the 36th International
  Conference on Machine Learning. vol.~97, pp. 6861--6871. PMLR (2019)

\bibitem{yang2019pointflow}
Yang, G., Huang, X., Hao, Z., Liu, M.Y., Belongie, S., Hariharan, B.:
  Pointflow: 3d point cloud generation with continuous normalizing flows. In:
  Proceedings of the IEEE International Conference on Computer Vision. pp.
  4541--4550 (2019)

\bibitem{shapenetpart}
Yi, L., Kim, V.G., Ceylan, D., Shen, I.C., Yan, M., Su, H., Lu, C., Huang, Q.,
  Sheffer, A., Guibas, L.: A scalable active framework for region annotation in
  3d shape collections. ACM Transactions on Graphics (ToG)  \textbf{35}(6),
  1--12 (2016)

\end{thebibliography}

\end{document}